%% file: 0-paper.tex
\documentclass[10pt,journal,compsoc]{IEEEtran}
\ifCLASSOPTIONcompsoc
  \usepackage[nocompress]{cite}
\else
  \usepackage{cite}
\fi
\ifCLASSINFOpdf
\else
\fi
\usepackage{comment}
\usepackage{ctable}
\usepackage{multirow}
\usepackage{booktabs} 
\usepackage{caption}
\usepackage{subcaption}
\usepackage{amsthm}
\usepackage{xspace}
\usepackage{mathtools}
\usepackage{enumitem}
\usepackage{xspace}
\usepackage{array}
\usepackage[group-separator={,}]{siunitx}
\usepackage{bm}
\usepackage{placeins}
\usepackage{soul}
\usepackage{multirow}
\usepackage{makecell}
\usepackage{booktabs}
\usepackage{enumitem}
\usepackage{multibib}
\usepackage{wrapfig}
\usepackage{makecell}
\usepackage{xcolor,colortbl}
\usepackage{dictsym}
\usepackage{csquotes}
\usepackage{amssymb}
\usepackage[pagebackref=true,colorlinks,bookmarks=false,citecolor=blue,linkcolor=blue,urlcolor=blue]{hyperref}
\def\eg{\emph{e.g}.} 
\def\ie{\emph{i.e}.}

\def\wrt{w.r.t. }

\begin{document}

\newcolumntype{L}[1]{>{\raggedright\let\newline\\\arraybackslash\hspace{0pt}}m{#1}}
\newcolumntype{X}[1]{>{\centering\let\newline\\\arraybackslash\hspace{0pt}}p{#1}}
\newcolumntype{Y}[1]{>{\raggedleft\let\newline\\\arraybackslash\hspace{0pt}}m{#1}}
\newcommand{\algname}{{ViDT}} 
\newtheorem{definition}{Definition}
\newtheorem{theorem}{Theorem}
\newtheorem{lemma}[theorem]{Lemma}
\definecolor{Gray}{gray}{0.90}

\title{An Extendable, Efficient and Effective Transformer-based Object Detector}
%
%
%
%

\author{Hwanjun Song, Deqing Sun, Sanghyuk Chun, Varun Jampani, Dongyoon Han,\\ Byeongho Heo, Wonjae Kim, Ming-Hsuan Yang
}

\IEEEtitleabstractindextext{%
\begin{abstract}
Transformers have been widely used in numerous vision problems especially for visual recognition and detection. 
Detection transformers are the first fully end-to-end learning systems for object detection, while vision transformers are the first fully transformer-based architecture for image classification.
In this paper, we integrate Vision and Detection Transformers\,(ViDT) to construct an effective and efficient object detector. 
ViDT introduces a reconfigured attention module to extend the recent Swin Transformer to be a standalone object detector,
followed by a computationally efficient transformer decoder that exploits multi-scale features and auxiliary techniques essential to boost the detection performance without much increase in computational load. 
{
In addition, we extend it to ViDT+ to support joint-task learning for object detection and instance segmentation. 
Specifically, we attach an efficient multi-scale feature fusion layer and utilize two more auxiliary training losses, IoU-aware loss and token labeling loss.
}
Extensive evaluation results on the Microsoft COCO benchmark dataset demonstrate that ViDT obtains the best AP and latency trade-off among existing fully transformer-based object detectors, {and its extended ViDT+ achieves {\bf 53.2}AP owing to its high scalability for large models.}
The source code and trained models are available at \url{https://github.com/naver-ai/vidt}.
\end{abstract}

}

\maketitle

\IEEEdisplaynontitleabstractindextext

%
\IEEEpeerreviewmaketitle

\input{1-introduction}

\input{2-methodology}

\input{3-evaluation}

\input{4-conclusion}


\bibliography{0-paper.bbl}

\vspace*{-1.5cm}
\begin{IEEEbiography}[{\includegraphics[width=1in,height=1.15in,clip,keepaspectratio]{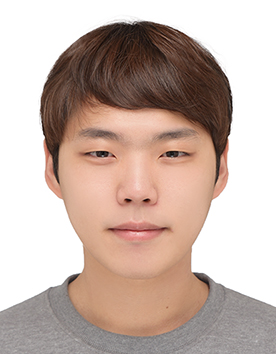}}]{Hwanjun Song} is a Research Scientist at the NAVER AI Lab. He worked as a Research Intern at Google Research in 2020 and received his Ph.D. degree in the Graduate School of Data Science from KAIST, Daejeon, Korea, in 2021. He is interested in designing advanced methodologies to handle data scale and quality issues, which are two main real-world challenges for AI. He was sponsored by Microsoft through Azure for Research from 2016 to 2018, and received the Qualcomm Innovation Award in 2019.
\end{IEEEbiography}
\vspace*{-1.1cm}

\begin{IEEEbiography}[{\includegraphics[width=1in,height=1.15in,clip,keepaspectratio]{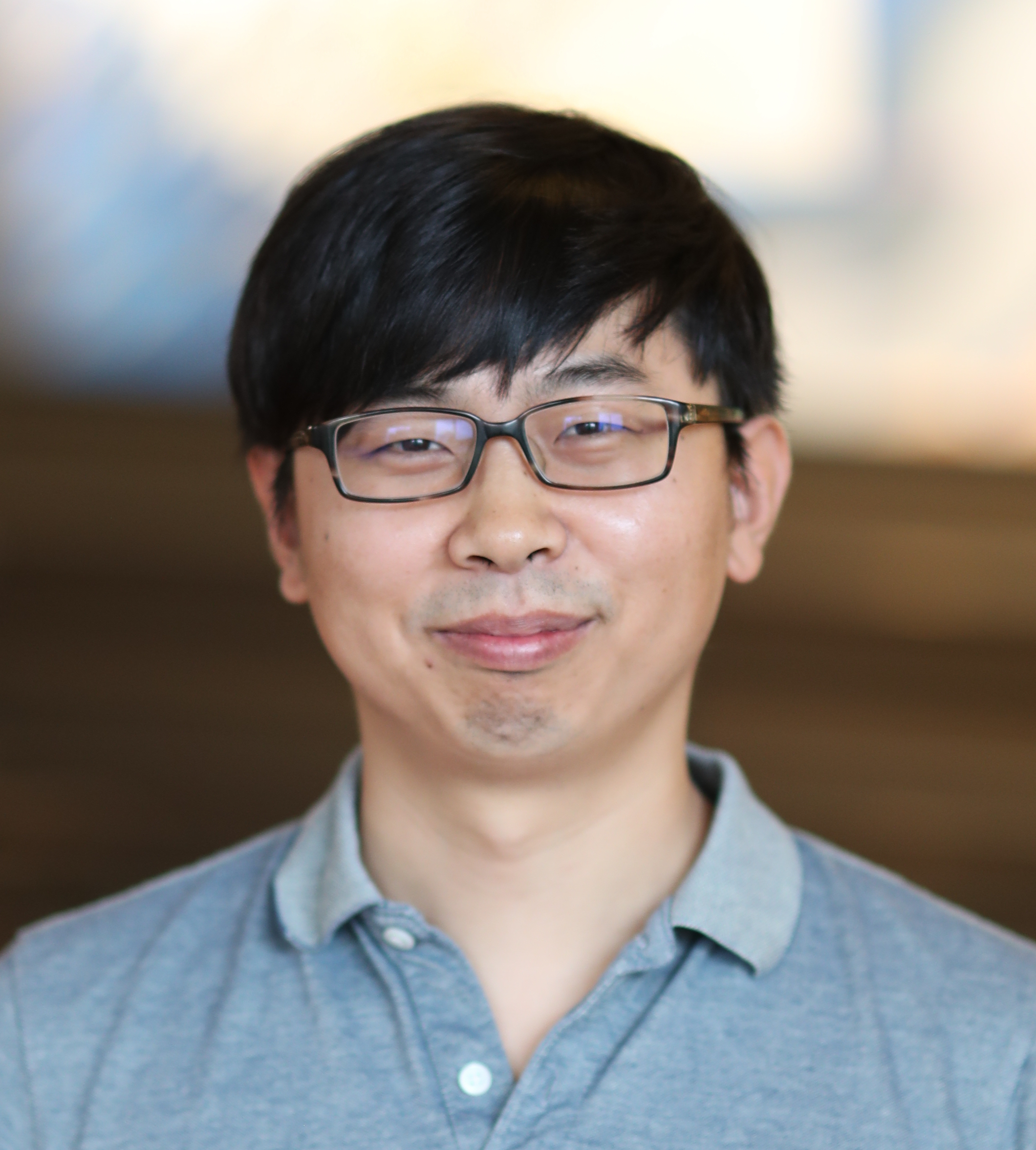}}]{Deqing Sun} is a staff research scientist and manager at Google working on computer vision and machine learning. He received a Ph.D. degree in Computer Science from Brown University. He served as an area chair for CVPR/ECCV/BMVC, and co-organized several workshops/tutorials at CVPR/ECCV/SIGGRAPH. He is a recipient of the best paper honorable mention award at CVPR 2018, the first prize in the robust optical flow competition at CVPR 2018 and ECCV 2020, the PAMI Young Researcher award in 2020, and the Longuet-Higgins prize at CVPR 2020.
\end{IEEEbiography}
\vspace*{-1.1cm}

\begin{IEEEbiography}[{\includegraphics[width=1in,height=1.15in,clip,keepaspectratio]{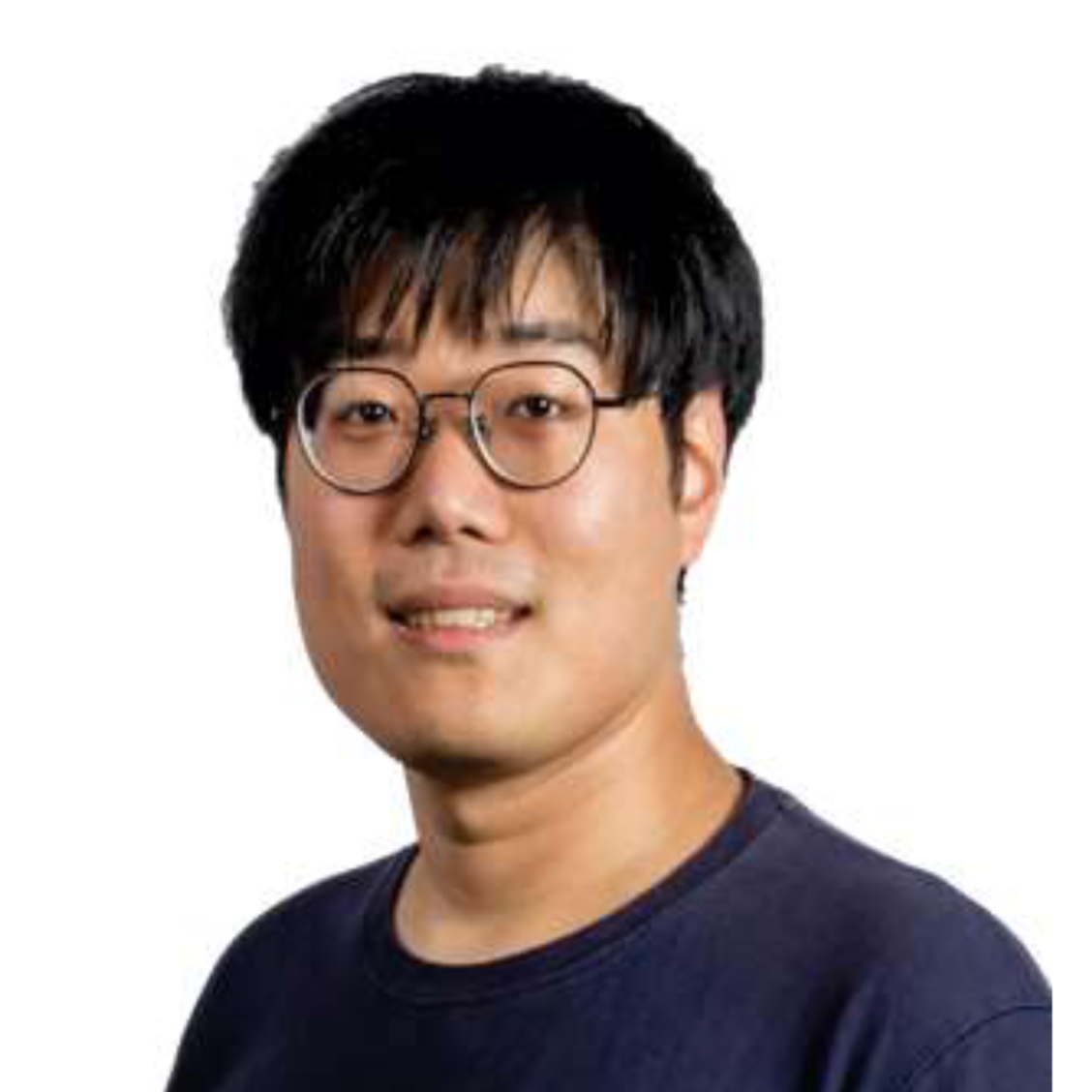}}]{Sanghyuk Chun} is a lead research scientist at the NAVER AI Lab. He was a research engineer at an advanced recommendation team in Kakao Corp from 2016 to 2018. He received his Master's and Bachelor's degrees in Electronical Engineering from KAIST, Daejeon, Korea, in 2016 and 2014, respectively. His research interests focus on reliable machine learning and vision-and-language.
\end{IEEEbiography}
\vspace*{-1.1cm}

\begin{IEEEbiography}[{\includegraphics[width=1in,height=1.15in,clip,keepaspectratio]{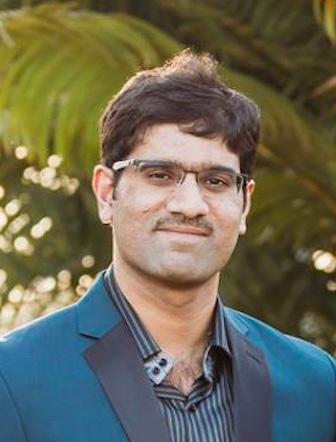}}]{Varun Jampani} is a researcher at Google Research working in the areas of machine learning and computer vision. His main research interests include self-supervised visual discovery, content-adaptive neural networks, and novel view synthesis. He obtained his PhD with highest honors at MPI for Intelligent Systems, Germany. He obtained his BTech and MS from IIIT-Hyderabad, India, where he was a gold medalist. He received Best Paper Honorable Mention award at CVPR2018. 
\end{IEEEbiography}
\vspace*{-1.1cm}

\begin{IEEEbiography}[{\includegraphics[width=1in,height=1.15in,clip,keepaspectratio]{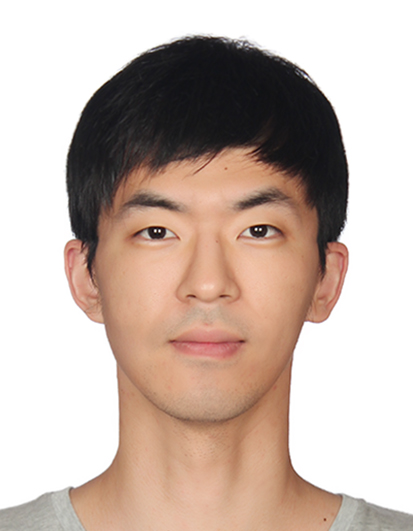}}]{Dongyoon Han} is a lead research scientist at NAVER AI Lab. His current research interests lie in machine learning and computer
vision, in particular, novel deep neural networks design and training methods. He received his B.S., M.S., and Ph.D. degrees in electrical engineering from Korea
Advanced Institute of Science and Technology,
Daejeon, South Korea, in 2011, 2013, and 2018,
respectively.
\end{IEEEbiography}
\vspace*{-1.1cm}

\begin{IEEEbiography}[{\includegraphics[width=1in,height=1.15in,clip,keepaspectratio]{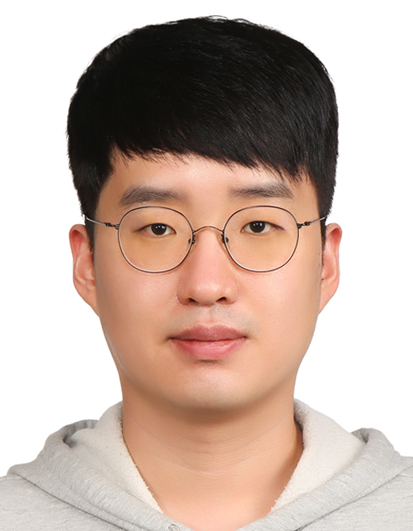}}]{Byeongho Heo} received the bachelor’s and Ph.D. degrees in electrical engineering and computer science from Seoul National University, Seoul, South Korea, in 2012 and 2019, respectively. In 2019, he joined the AI LAB, NAVER Corporation, as a Research Scientist, where he is currently working. His current research interests include vision transformer, knowledge distillation, image classification, and optimizer for deep learning.
\end{IEEEbiography}

\vspace*{-1.1cm}
\begin{IEEEbiography}[{\includegraphics[width=1in,height=1.15in,clip,keepaspectratio]{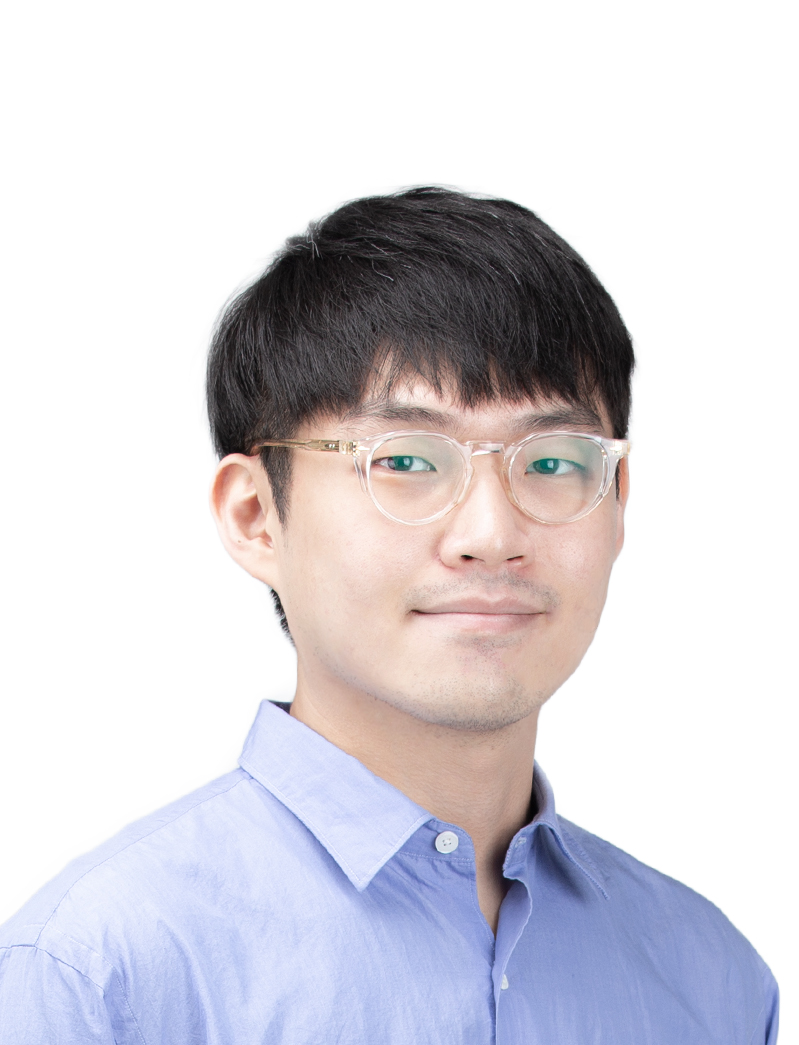}}]{Wonjae Kim} is a research scientist at NAVER AI Lab. Before joining NAVER, He worked as a research scientist at Kakao corporation from 2018 to 2021. He received his Master's and Bachelor's degrees in computer science and engineering from Seoul national university, in 2018 and 2016, respectively. His research interests include vision-and-language representation learning, human-computer interaction, and information visualization.
\end{IEEEbiography}

\vspace*{-1.1cm}
\begin{IEEEbiography}[{\includegraphics[width=1in,height=1.15in,clip,keepaspectratio]{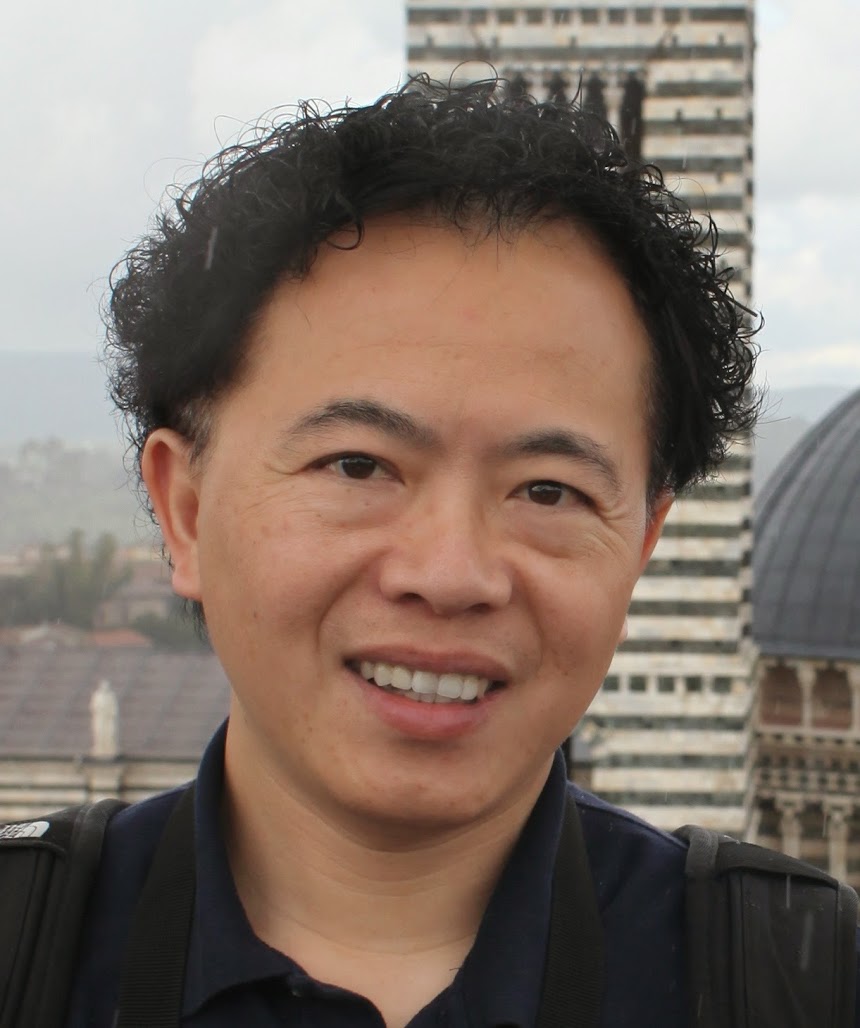}}]{Ming-Hsuan Yang}
is a professor of Electrical Engineering and Computer Science with the University of California, Merced, CA, USA.
He received the Ph.D. degree in computer science from the University of Illinois at Urbana-Champaign, USA, in 2000.
He served as an Associate Editor of the IEEE Transactions on Pattern Analysis and Machine Intelligence from 2007 to 2011,
and is an Associate Editor of the International Journal of Computer Vision, and Image and Vision Computing.
He received the NSF CAREER Award in 2012, and Google Faculty Award in 2009. He is a fellow of the IEEE and the ACM..
\end{IEEEbiography}

\clearpage
\input{6-appendix}

\end{document}

%% file: 1-introduction.tex
\section{Introduction}
\label{sec:introduction}

\IEEEPARstart{O}{bject} detection aims to predict both the bounding box and object class for each object of interest in an image.
Recent deep object detectors rely heavily on meticulously designed components, such as anchor generation and non-maximum suppression\,\cite{papageorgiou2000trainable, liu2020deep}.
As a result, the performance of these object detectors depends on specific postprocessing steps, which involve complex pipelines and make fully end-to-end training difficult.

Motivated by the recent success of Transformers\,\cite{vaswani2017attention} in NLP, numerous models 
have been developed for various vision tasks, especially in recognition and detection. 
Carion et al.\,\cite{carion2020end} propose the Detection Transformers\,(DETR) to replace the meticulously designed components with a transformer encoder and decoder architecture, which serves as a neck component to bridge a CNN body for feature extraction and a detector head for prediction.
As such, DETR enables end-to-end training of deep object detectors. 
On the other hand, Dosovitskiy et al.\,\cite{dosovitskiy2021image} show that a fully-transformer
backbone without any convolutional layers, Vision Transformer (ViT), achieves state-of-the-art results on image classification benchmarks.
%
DETR and ViT have been shown to learn effective representation models without relying strongly on human inductive biases, \eg, meticulously designed components in object detection\,(DETR), convolutional layers and pooling mechanisms for locality-aware designs\,(ViT). 
However, no attempt has been made to synergize DETR and ViT for a better object detection architecture.
In this work, we integrate both approaches to construct a fully transformer-based, end-to-end object detector that achieves state-of-the-art performance without increasing computational load. 

\begin{figure}[t!]
\includegraphics[width=.95\columnwidth]{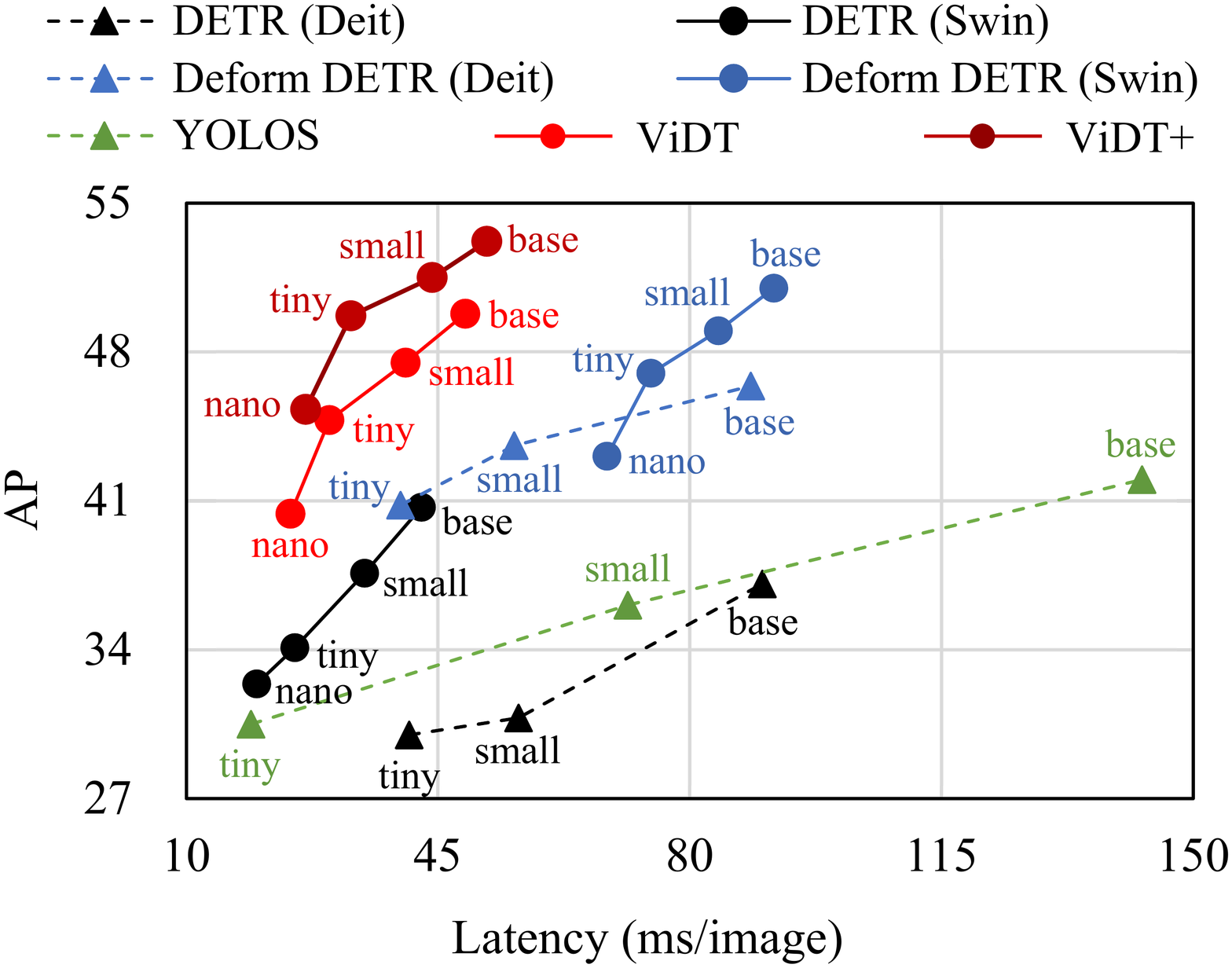}
\vspace*{-0.15cm}
\caption{{Performance of recent object detectors in terms of average precision (AP) and latency. Detailed  The text in the plot indicates the backbone model size. The latency was measured with batch size 1 of $800 \times 1333$ resolution on NVIDIA A100 GPU.} \color{black}AP and latency\,(milliseconds) are summarized in Table \ref{table:full_exp_coco}.}
\label{fig:ap_vs_latency}
\vspace*{-0.4cm}
\end{figure}

\begin{figure*}[t!]
\begin{center}
\includegraphics[width=.95\textwidth]{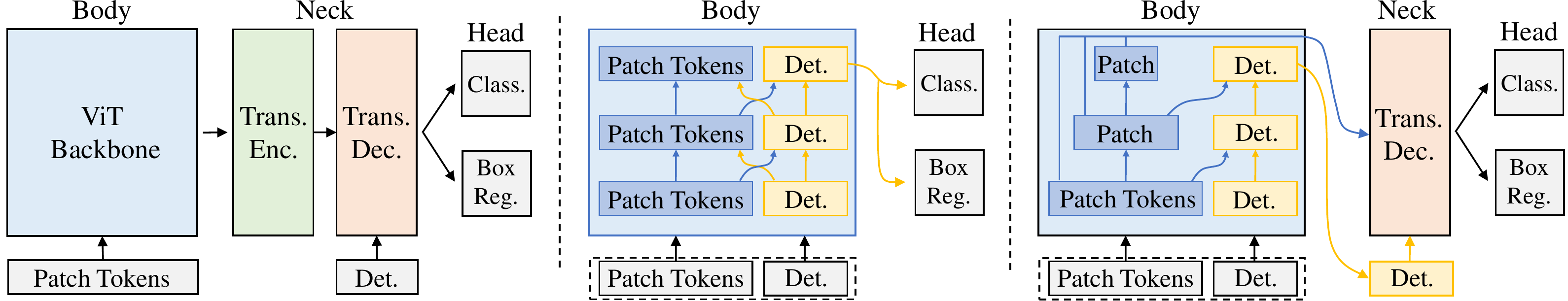}
\end{center}
\vspace*{-0.1cm}
\hspace*{2.2cm} {\small (a) DETR\,(ViT).} \hspace*{3.9cm} {\small (b) YOLOS.} \hspace*{3.5cm} {\small \textbf{ (c) \algname{} (Ours)}.}
\vspace*{-0.1cm}
\caption{Pipelines of fully transformer-based object detectors. DETR (ViT) denotes Detection Transformer using ViT as its body.
\algname{} exploits merits of DETR\,(ViT) and YOLOS and achieves the best AP and latency trade-off among fully transformer-based object detectors. 
}
\label{fig:pipelines}
\vspace*{-0.4cm}
\end{figure*}

Straightforward integration of DETR and ViT can be achieved by replacing the ResNet backbone (body) of DETR with ViT as shown in Figure \ref{fig:pipelines}(a).
This naive integration, {DETR\,(ViT)}\footnote{We refer to each model based on the combinations of its body and neck. For example, DETR\,(DeiT) indicates that DeiT\,(vision transformers) is integrated with DETR\,(detection transformers).}, has two limitations. 
First, as the original ViT suffers from the quadratic increase in complexity \wrt image size, this approach does not scale up well. 
Furthermore, the attention operation at the transformer encoder and decoder (\ie, the ``neck'' component) adds significant computational overhead to the detector. 
Thus, the naive integration of DETR and ViT would cause high latency, as shown in the blue lines of Figure \ref{fig:ap_vs_latency}.

Recently, Fang et al.\,\cite{fang2021you} propose the \textbf{YOLOS} model 
by appending the detection tokens $[\mathtt{DET}]$ to the patch tokens $[\mathtt{PATCH}]$\, (see Figure \ref{fig:pipelines}(b)), where $[\mathtt{DET}]$ tokens are learnable embeddings to specify different objects to detect.
YOLOS is a neck-free architecture and removes the additional computational costs from the neck encoder.
However, YOLOS shows limited performance because it cannot exploit additional optimization techniques on the neck architecture, \eg, multi-scale features and auxiliary loss. 
In addition, YOLOS can only accommodate the original transformer due to its architectural limitation, resulting in a quadratic complexity \wrt the input size.\looseness=-1

\begin{figure*}[t!]
\begin{center}
\includegraphics[width=1.0\textwidth]{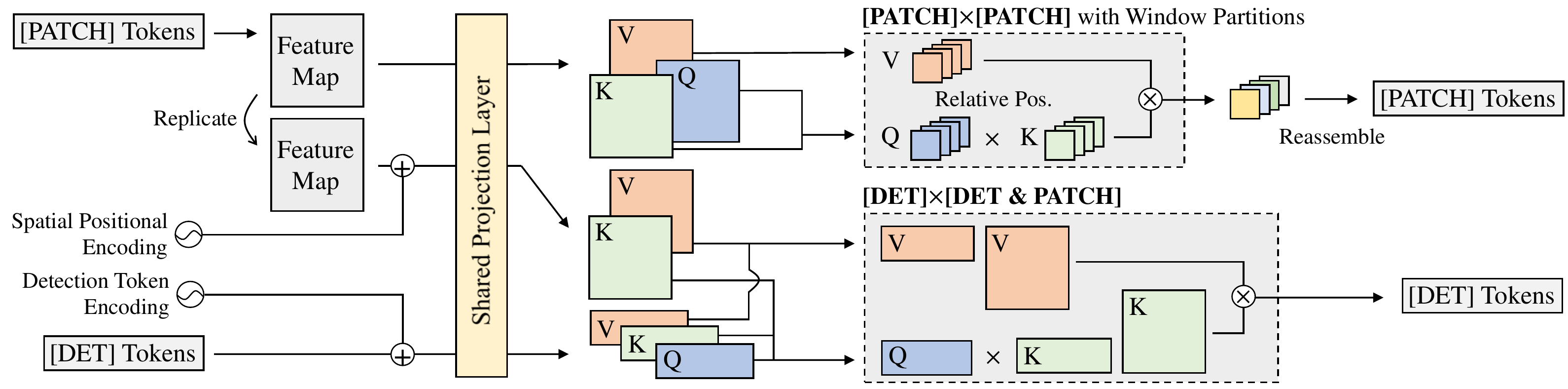}
\end{center}
\vspace*{-0.2cm}
\caption{Reconfigured Attention Module (Q: query, K: key, V: value). The skip connection and feedforward networks following the attention operation is omitted just for ease of exposition.}
\label{fig:reconfigured_attention}
\vspace*{-0.3cm}
\end{figure*}

In this paper, we propose a novel integration of {\underline{Vi}}sion and {\underline{D}}etection {\underline{T}}ransformers\,({\algname{}})\,(see Figure \ref{fig:pipelines}(c)).
Our contributions are three-fold.
{First}, ViDT introduces the Reconfigured Attention Module\,(RAM), to facilitate any ViT variant to handle the appended $[\mathtt{DET}]$ and $[\mathtt{PATCH}]$ tokens for object detection. 
Thus, we can integrate the Swin Transformer~\cite{liu2021swin} backbone with RAM to be an object detector and obtain high scalability using its local attention mechanism with linear complexity.
{Second}, ViDT adopts a lightweight encoder-free neck architecture to reduce the computational overhead 
while still enabling the additional optimization techniques on the neck module. 
Note that the neck encoder is unnecessary because RAM directly extracts fine-grained representation for object detection, \ie, $[\mathtt{DET]}$ tokens.
As a result, ViDT achieves better performance than its neck-free counterparts. 
{\color{black}
Finally, we extend the vanilla ViDT to an end-to-end architecture named {ViDT+}, thereby enabling multi-task learning of object detection and instance segmentation. 
For effective multi-task learning, ViDT+ equips efficient feature pyramid layers on top of its body for multi-scale feature fusion, and leverages two additional training losses \emph{i.e.}, IoU-aware loss\,\cite{wu2020iou} and token labeling loss\,\cite{jiang2021all}. %
ViDT+ only adds 1M learnable parameters and barely reduces its inference speed than ViDT, but achieves a significant accuracy gain.
}

ViDT has {\color{black}three} architectural advantages over existing approaches. 
First, similar to YOLOS, ViDT takes $[\mathtt{DET}]$ tokens as the additional input, maintaining a fixed scale for object detection, but constructs hierarchical representations starting with small-sized image patches for $[\mathtt{PATCH}]$ tokens.
Second, ViDT can use the hierarchical\,(multi-scale) features and additional techniques without a significant computation overhead.  Thus, as a fully transformer-based object detector, ViDT facilitates better integration of vision and detection transformers. 
{\color{black}
Third, ViDT accommodates numerous tasks and transformer models. 
It can be extended to be an end-to-end architecture for multi-task learning, and easily combined with ViT variants such as CoaT\,\cite{xu2021co} and PVT-v2\,\cite{wang2021pvtv2} other than the Swin Transformer.
}
Extensive experiments on Microsoft COCO~\cite{lin2014microsoft} show that ViDT is highly scalable even for large ViT models, such as Swin-base with 0.1 billion parameters, and achieves the best AP and latency trade-off among existing fully transformer-based detectors. 
{\color{black}
In particular, ViDT+ with Swin-base achieves 53.2AP for object detection, which is $2.6$AP higher than that of the vanilla ViDT.
}

This work is an extensive version of our ICLR 2022\,\cite{song2022vidt}. Compared to the ICLR 2022 work, this paper includes the following additional contributions: (1) a new joint-learning framework named ViDT+, an extension of the vanilla ViDT, by incorporating three additional components: an efficient pyramid feature fusion module, a unified query representation module, and an IoU-aware and token labeling losses, (2) a detailed computational complexity analysis of the proposed VIDT compared to the YOLOS detector, (3) a performance comparison for object detection and instance segmentation compared with other CNN-based state-of-the-art methods, (4) an ablation study for the three new components of ViDT+ and a complete analysis of all the proposed components, (5) the reconfigured attention module combined with ViT variants such as CoaT and PVT-v2 other than the Swin Transformer.

\section{Preliminaries}
\label{sec:preliminary}

\textbf{Object detection}\, is a task of predicting a set of bounding boxes and classification labels for each object of interest. There have been significant efforts to develop efficient and effective detection backbones and pipelines\,\cite{li2022exploring, ge2021yolox}. Typically, as a pre-trained backbone is fine-tuned in single- and two-stage fashion; single-stage detectors produce predictions directly w.r.t. anchors or a grid of possible object centers, \emph{e.g.}, SSD\,\cite{liu2016ssd} and YOLO\,\cite{redmon2016you}, while two-stage detectors predict bounding boxes w.r.t. region proposals, \emph{e.g.},. Faster R-CNN\,\cite{ren2015faster}. The success of modern object detectors has achieved very high detection accuracy, but they heavily depend on some meticulously designed components, such as anchor generation and non-maximum suppression\,\cite{papageorgiou2000trainable, liu2020deep}. In addition, finding the optimal trade-off between detection accuracy and inference speed is not a trivial task. In this paper, we study the integration of vision and detection transformers and show its potential to be a new generic detection pipeline, achieving good performance trade-offs without the meticulously designed components.

\smallskip\smallskip
\noindent\textbf{Vision transformers}\, process an image as a sequence of small image patches, thereby facilitating consideration of 
interaction among patches at all positions \,(\ie, global attention). 
However, the original ViT\,\cite{dosovitskiy2021image} cannot be easily scaled to a wide range of vision tasks due to its high computational complexity, which increases quadratically with respect to image size. 
The Swin Transformer\,\cite{liu2021swin} alleviates the complexity issue by introducing the notion of {shifted windows} that support {local attention} and {patch reduction} operations, thereby improving compatibility for dense prediction tasks such as object detection and semantic segmentation.
A few approaches use vision transformers as detector backbones but achieve limited success\,\cite{liu2021swin,heo2021rethinking,fang2021you}.
In this work, we significantly improve the performance of detectors with transformer backbones by the proposed architectural changes {\color{black}and extensions for multi-task learning.}

\smallskip\smallskip
\noindent\textbf{Detection transformers} \, eliminate the meticulously designed components
(\eg, anchor generation and non-maximum suppression)
by combining convolutional network backbones and Transformer encoder-decoders. 
While the original DETR\,\cite{carion2020end} achieves high detection performance, it suffers from slow convergence compared to previous detectors. 
For example, DETR requires 500 epochs while the Faster R-CNN\,\cite{ren2015faster} needs only 37 epochs\,\cite{wu2019detectron2} for training. 
To mitigate the issue, Zhu et al.\,\cite{zhu2021deformable} propose Deformable DETR which introduces deformable attention for utilizing multi-scale features as well as expediting the slow training process of DETR.
In this paper, we use the Deformable DETR as our base detection transformer framework and integrate it with the {\color{black}three} recent vision transformers.

\smallskip\smallskip
\noindent\textbf{DETR\,(ViT)} \, is a straightforward integration of DETR and ViT, which uses ViT as a feature extractor, followed by the transformer encoder-decoder in DETR. 
As illustrated in Figure \ref{fig:pipelines}(a), it is a \emph{body--neck--head} structure; the representation of input $[\mathtt{PATCH}]$ tokens are extracted by the ViT backbone and then directly fed to the transformer-based encoding and decoding pipeline. 
To predict multiple objects, a fixed number of learnable $[\mathtt{DET}]$ tokens are provided as additional input to the decoder. 
Subsequently, the output embeddings by the decoder are considered as final predictions through the detection heads for classification and box regression.
Since DETR\,(ViT) does not modify the backbone at all, it can be flexibly changed to any latest ViT model, \eg, Swin Transformer. 
Additionally, its neck decoder facilitates the aggregation of multi-scale features and the use of additional techniques, such as auxiliary decoding loss and iterative box refinement, which help detect objects of different sizes and speed up the  training process\,\cite{zhu2021deformable}.
However, the attention operation at the neck encoder adds significant computational overhead to the detector.
In contrast, \algname{} resolves this issue by directly extracting fine-grained $[\mathtt{DET}]$ features from the Swin Transformer with RAM without maintaining the transformer encoder in the neck architecture.

\smallskip\smallskip
\noindent\textbf{YOLOS} \cite{fang2021you} \, is a ViT architecture for object detection with minimal modifications. 
As illustrated in Figure \ref{fig:pipelines}(b), YOLOS consists of a \emph{neck-free} structure by appending randomly initialized learnable $[\mathtt{DET}]$ tokens to the sequence of input $[\mathtt{PATCH}]$ tokens.
Since all the embeddings for $[\mathtt{PATCH}]$ and $[\mathtt{DET}]$ tokens interact via global attention, the final $[\mathtt{DET}]$ tokens are generated by the fine-tuned ViT backbone and then directly generate predictions through the detection heads without requiring any neck layer.
While the naive DETR (ViT) suffers from the computational overhead from the neck layer, YOLOS enjoys efficient computations
by treating the $[\mathtt{DET}]$ tokens as additional input for ViT. 
YOLOS shows that 2D object detection can be accomplished in a pure sequence-to-sequence manner, but this solution entails \emph{two} inherent limitations:
{
\renewcommand\labelenumi{(\theenumi)}
\begin{enumerate}[leftmargin=15pt]
\item YOLOS inherits the drawback of the original ViT; the {high computational complexity} attributed to the global attention operation. 
As illustrated in Figure \ref{fig:ap_vs_latency}, YOLOS shows poor latency compared with other fully transformer-based detectors, especially when its model size becomes larger, \ie, small $\rightarrow$ base. Thus, YOLOS is \emph{not scalable} for the large model.
\vspace*{0.1cm}
\item YOLOS does not benefit from using any additional techniques essential for better detection performance, \eg, multi-scale features, due to the absence of the neck layer.
Although YOLOS used the same DeiT backbone with Deformable DETR\,(DeiT), its AP was lower than the straightforward integration. 
\end{enumerate}
}
In contrast, the encoder-free neck architecture of ViDT enjoys  additional optimization techniques from Zhu et al.\,\cite{zhu2021deformable}, resulting in faster convergence and better performance.
Further, our RAM enables us to combine the Swin Transformer\footnote{{\color{black}It is not limited to the Swin Transformer. 
Our reconfiguration scheme can be easily applied to other variants with simple modifications. See Appendix \ref{appendix:ram_other_vit} for the combination with CoaT and PVT-v2.}} and sequence-to-sequence paradigm for detection. 
%

%% file: 2-methodology.tex
\section{ViDT: Vision and Detection Transformers\!\!\!\!\!}
\label{sec:method}

\algname{} first reconfigures the attention model of the  Swin Transformer to support standalone object detection while fully reusing the parameters of the Swin Transformer. 
Next, it incorporates an encoder-free neck layer to exploit multi-scale features and two essential techniques: auxiliary decoding loss and iterative box refinement. 

\subsection{Reconfigured Attention Module\,(RAM)}

Applying patch reduction and local attention scheme
of the Swin Transformer to the sequence-to-sequence paradigm is challenging because (1)\,the number of $[\mathtt{DET}]$ tokens must be maintained at a fixed-scale and (2)\,the lack of locality between $[\mathtt{DET}]$ tokens. 
To address these issues, we introduce a reconfigured attention module\,(RAM) that decomposes a single global attention associated with $[\mathtt{PATCH}]$ and $[\mathtt{DET}]$ tokens into the \emph{three} different attention, namely $[\mathtt{PATCH}]\times[\mathtt{PATCH}]$, $[\mathtt{DET}]\times[\mathtt{DET}]$, and $[\mathtt{DET}]\times[\mathtt{PATCH}]$ attention.
Based on the decomposition, the efficient schemes of the Swin Transformer are applied only to $[\mathtt{PATCH}]\times[\mathtt{PATCH}]$ attention, which is the heaviest part of computational complexity, without breaking the two constraints on $[\mathtt{DET}]$ tokens. 
As illustrated in Figure \ref{fig:reconfigured_attention}, these modifications fully reuse all the parameters of the Swin Transformer by sharing projection layers for $[\mathtt{DET}]$ and $[\mathtt{PATCH}]$ tokens, and perform the three different attention operations:

As a standalone object detector, RAM must be accompanied by three attention operations:
\begin{itemize}[leftmargin=10pt]
\item {$[\mathtt{PATCH}]\times[\mathtt{PATCH}]$ Attention:} The initial $[\mathtt{PATCH}]$ tokens are progressively calibrated across the attention layers such that they aggregate the key contents in the global feature map\,(\ie, a spatial form of $[\mathtt{PATCH}]$ tokens) according to the attention weights, which are computed by $\langle$query, key$\rangle$ pairs. 
For $[\mathtt{PATCH}]\times[\mathtt{PATCH}]$ attention, the Swin Transformer performs local attention on each window partition, but its shifted window partitioning in successive blocks bridges the windows of the preceding layer, providing connections among partitions to capture global information. 
We use a similar policy to generate hierarchical $[\mathtt{PATCH}]$ tokens. 
Thus, the number of $[\mathtt{PATCH}]$ tokens is 
reduced by a factor of 4 at each stage; the resolution of feature maps decreases from {\small $H/4 \times W/4$} to {\small $H/32 \times W/32$} over a total of four stages, where $H$ and $W$ denote the width and height of the input image.
\vspace*{0.1cm}
\item {$[\mathtt{DET}]\times[\mathtt{DET}]$ Attention:} Similar to YOLOS, we append one hundred learnable $[\mathtt{DET}]$ tokens as the additional input to the Swin Transformer.
As the number of $[\mathtt{DET}]$ tokens specify the number of objects to detect, their number must be maintained with a fixed-scale over the transformer layers. 
In addition, $[\mathtt{DET}]$ tokens do not have any locality unlike the $[\mathtt{PATCH}]$ tokens. 
Hence, for $[\mathtt{DET}]\times[\mathtt{DET}]$ attention, we perform global self-attention while maintaining the number of them; this attention helps each $[\mathtt{DET}]$ token to localize a different object by capturing the relationship between them.
\vspace*{0.1cm}
\item {$[\mathtt{DET}]\times[\mathtt{PATCH}]$ Attention:} We consider cross-attention between $[\mathtt{DET}]$ and $[\mathtt{PATCH}]$ tokens, and generate an object embedding per $[\mathtt{DET}]$ token. 
For each $[\mathtt{DET}]$ token, the key contents in $[\mathtt{PATCH}]$ tokens are aggregated to represent the target object. 
Since the $[\mathtt{DET}]$ tokens specify different objects, it produces different object embeddings for diverse objects in the image.
Without the cross-attention, it is infeasible to realize the standalone object detector. 
As shown in Figure \ref{fig:reconfigured_attention}, \algname{} binds $[\mathtt{DET}]\times[\mathtt{DET}]$ and $[\mathtt{DET}]\times[\mathtt{PATCH}]$ attention to process them at once to increase efficiency.
\end{itemize}
We replace all the attention modules in the Swin Transformer with the proposed RAM, which receives $[\mathtt{PATCH}]$ and $[\mathtt{DET}]$ tokens (as shown in ``Body'' of Figure \ref{fig:pipelines}(c)) and then outputs their calibrated new tokens by performing the three different attention operations in parallel.

\smallskip\smallskip
\noindent\textbf{Positional Encoding.}  \algname{} adopts different positional encodings for different types of attention. 
For $[\mathtt{PATCH}]\times[\mathtt{PATCH}]$ attention, we use the relative position bias\,\cite{hu2019local} originally used in the Swin Transformer. 
In contrast, the learnable positional encoding is added for $[\mathtt{DET}]$ tokens for $[\mathtt{DET}]\times[\mathtt{DET}]$ attention because there is no particular order between $[\mathtt{DET}]$ tokens. 
However, for $[\mathtt{DET}] \times[\mathtt{PATCH}]$ attention, it is crucial to inject \emph{spatial bias} to the $[\mathtt{PATCH}]$ tokens due to the permutation-equivariant in transformers, ignoring spatial information of the feature map.
Thus, \algname{} adds the sinusoidal-based spatial positional encoding to the feature map, which is reconstructed from the $[\mathtt{PATCH}]$ tokens for $[\mathtt{DET}]\times[\mathtt{PATCH}]$ attention, as can be seen from the left side of Figure \ref{fig:reconfigured_attention}. 
We present a thorough analysis of various spatial positional encodings in Section \ref{sec:exp_ram}. 

\smallskip\smallskip
\noindent\textbf{Use of $\bm{[\mathtt{DET}]\times[\mathtt{PATCH}]}$ Attention.} Applying cross-attention between $[\mathtt{DET}]$ and $[\mathtt{PATCH}]$ tokens adds additional computational overhead to the Swin Transformer, especially when it is activated at the bottom layer due to the large number of $[\mathtt{PATCH}]$ tokens. 
To minimize such computational overhead, \algname{} only activates 
the cross-attention at the last stage\,(the top level of the pyramid) of the Swin Transformer,
which consists of two transformer layers that receives $[\mathtt{PATCH}]$ tokens of size {\small $H/32 \times W/32$}. 
Thus, only self-attention for $[\mathtt{DET}]$ and $[\mathtt{PATCH}]$ tokens are performed for the remaining stages except the last one.  
In Section \ref{sec:exp_ram}, we show that this design choice helps achieve the highest FPS, while achieving similar detection performance as when cross-attention is enabled at every stage.

\begin{table*}[t!]
\begin{center}
\caption{{\color{black}Summary of computational complexity for different attention operations used in YOLOS and ViDT (RAM), where ${\sf P}$ and ${\sf D}$ are the number of $[\mathtt{PATCH}]$ and $[\mathtt{DET}]$ tokens, respectively (${\sf D} \ll {\sf P}$).}}
\label{table:full_analysis}
\vspace*{-2mm}
\small
\begin{tabular}{L{4.0cm} |X{6.45cm} |X{6.45cm}}\hline
\rowcolor{Gray}
Attention Type     \!\! & YOLOS & ViDT \\\toprule
$[\mathtt{PATCH}]\times[\mathtt{PATCH}]$ Attention\!\! & $\mathcal{O}(d^2 {\sf P} + d {\sf P}^2)$  &  $\mathcal{O}(d^2 {\sf P} + d k^2 {\sf P})$ \\
$[\mathtt{DET}]\times[\mathtt{DET}]$\!\! Attention & $\mathcal{O}(d^2 {\sf D} + d{\sf D}^{2})$  &  $\mathcal{O}(d^2 {\sf D} + d {\sf D}^{2})$ \\
$[\mathtt{DET}]\times[\mathtt{PATCH}]$\!\! Attention & $\mathcal{O}(d{\sf P}{\sf D})$ &  $\mathcal{O}(d^2 ({\sf P} + {\sf D}) + d {\sf P} {\sf D})$ \\\midrule
Total Complexity\!\! & $\mathcal{O}(d^2 ({\sf P} + {\sf D}) + $ $d ({\sf P} + {\sf D})^{2})$ & $\mathcal{O}(d^2 ({\sf P} + {\sf D}) + dk^2 {\sf P} + d {\sf D}^{2} + d {\sf P} {\sf D})$ \\\bottomrule
\end{tabular}
\end{center}
\vspace*{-4mm}
\end{table*}

\smallskip\smallskip
{\color{black}
\noindent\textbf{Binding $\bm{[\mathtt{DET}]\times[\mathtt{DET}]}$ and $\bm{[\mathtt{DET}]\times[\mathtt{PATCH}]}$ Attention.} 
We bind the two attention modules by a simple implementation.
In this work, $[\mathtt{DET}]\times[\mathtt{DET}]$ and $[\mathtt{DET}]\times[\mathtt{PATCH}]$ attention generate new $[\mathtt{DET}]$ tokens, which aggregate relevant contents in $[\mathtt{DET}]$ and $[\mathtt{PATCH}]$ tokens, respectively. 
Since the two attention modules share exactly the same $[\mathtt{DET}]$ query embedding obtained after the projection as shown in Figure \ref{fig:reconfigured_attention}, they can be processed at once by performing matrix multiplication between $[\mathtt{DET}]_{Q}$ and $\big[ ~[\mathtt{DET}]_{K} \cdot [\mathtt{PATCH}]_{K}\big]$ embeddings, where $Q$, $K$ are the key and query, and $[\cdot]$ is the concatenation. 
Then, the obtained attention map is applied to the $\big[ ~[\mathtt{DET}]_{V} \cdot [\mathtt{PATCH}]_{V}\big]$ embeddings, where $V$ is the value and $d$ is the embedding dimension,
\begin{equation}
\begin{gathered}
[\mathtt{DET}]_{new} = \\ \text{Softmax}\big(\frac{[\mathtt{DET}]_{Q} \big[[\mathtt{DET}]_{K},[\mathtt{PATCH}]_{K}\big]^{\top}}{\sqrt{d}}) \big[[\mathtt{DET}]_{V}, [\mathtt{PATCH}]_{V}\big].
\end{gathered}
\end{equation}

\smallskip\smallskip
\noindent\textbf{Embedding Dimension of $[\mathtt{DET}]$ tokens.}
In this work, $[\mathtt{DET}]\times[\mathtt{DET}]$ attention is performed across every stage, and the embedding dimension of $[\mathtt{DET}]$ tokens increases gradually like $[\mathtt{PATCH}]$ tokens.
For a $[\mathtt{PATCH}]$ token, its embedding dimension is increased by concatenating nearby $[\mathtt{PATCH}]$ tokens in a grid. 
However, this mechanism does not apply to $[\mathtt{DET}]$ tokens since we maintain the same number of $[\mathtt{DET}]$ tokens for detecting a fixed number of objects in a scene. 
Hence, we simply repeat a $[\mathtt{DET}]$ token multiple times along the embedding dimension to increase its size. 
This allows $[\mathtt{DET}]$ tokens to reuse all the projection and normalization layers in the Swin Transformer without any modification\footnote{\color{black} When combined RAM with CoaT and PVT-v2, which use convolutional layers for token pooling, we add linear layers to handle the embedding dimension of $[\mathtt{DET}]$ tokens.}.
}



\subsection{Encoder-free Neck Structure}

To exploit multi-scale feature maps, \algname{} incorporates a decoder of multi-layer deformable transformers\,\cite{zhu2021deformable}. 
In the DETR family (Figure \ref{fig:pipelines}(a)), a transformer encoder is required at the neck to transform features extracted from the backbone for image classification into the ones suitable for object detection; the encoder is generally computationally expensive since it involves $[\mathtt{PATCH}]\times[\mathtt{PATCH}]$ attention. 
However, \algname{} maintains only a transformer decoder as its neck, in that the Swin Transformer with RAM directly extracts fine-grained features suitable for object detection as a standalone object detector. 
Thus, the neck structure of \algname{} is computationally efficient. 

The decoder receives two inputs from the Swin Transformer with RAM: (1) $[\mathtt{PATCH}]$ tokens generated from each stage\,(\ie, four multi-scale feature maps, $\{{\bm x^{l}}\}_{l=1}^{L}$ where $L=4$) and (2) $[\mathtt{DET}]$ tokens generated from the last stage. The overview is illustrated in ``Neck'' of Figure \ref{fig:pipelines}(c).
In each deformable transformer layer, $[\mathtt{DET}]\times[\mathtt{DET}]$ attention is performed first. 
For each $[\mathtt{DET}]$ token, multi-scale deformable attention is applied to produce a new $[\mathtt{DET}]$ token, aggregating a small set of key contents sampled from the multi-scale feature maps $\{{\bm x^{l}}\}_{l=1}^{L}$,
\begin{equation}
\label{eq:msdeform_attn}
\begin{gathered}
{\rm MSDeformAttn}([\mathtt{DET}], \{ {\bm x^{l}} \}_{l=1}^{L}) =\\ \sum_{m=1}^{M} {\bm W_m} \Big[ \sum_{l=1}^{L} \sum_{k=1}^{K} A_{mlk} \cdot {\bm W_m^{\prime}} {\bm x^{l}}\big(\phi_{l}({\bm p}) + \Delta {\bm p_{mlk}}\big) \Big],\\
\end{gathered}
\end{equation}
where $m$ indices the attention head and $K$ is the total number of sampled keys for content aggregation. 
In addition, $\phi_{l}({\bm p})$ is the reference point of the $[\mathtt{DET}]$ token re-scaled for the $l$-th level feature map, while $\Delta{\bm p_{mlk}}$ is the sampling offset for deformable attention;  and 
$A_{mlk}$ is the attention weights of the $K$ sampled contents. ${\bm W_m}$ and ${\bm W_m^{\prime}}$ are the projection matrices for multi-head attention.

\begin{figure*}[t!]
\begin{center}
\includegraphics[width=18cm]{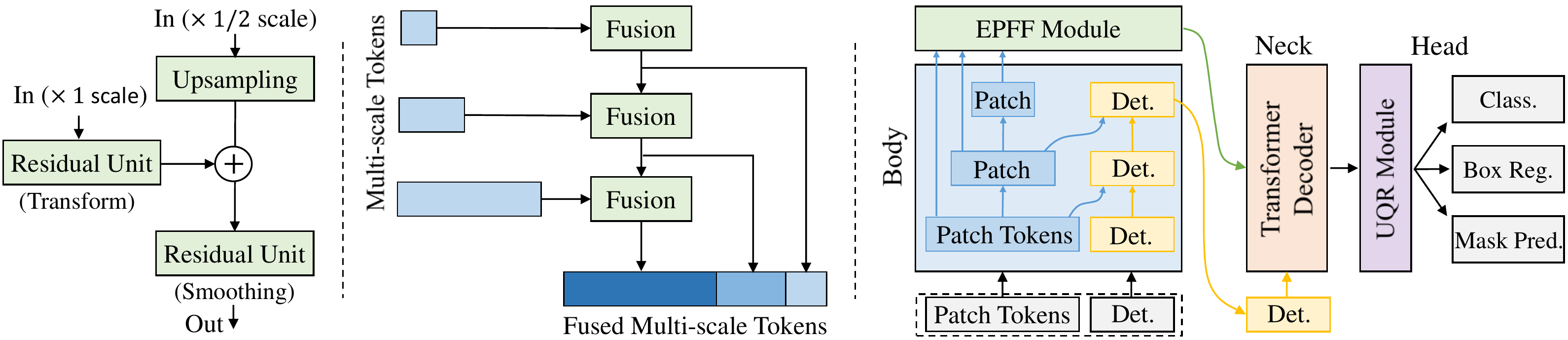}
\end{center}
\vspace*{-0.1cm}
\hspace*{0.8cm} {\small (a) Fusion Block.} \hspace*{2.25cm} {\small (b) EPFF Module.} \hspace*{5.35cm} {\small { (c) ViDT+}.}
\vspace*{-0.1cm}
\caption{{\color{black}Architecture overview of ViDT+: (a)\,two inputs with different scales are fused by a fusion block, (b)\,multiple fusion blocks are used in the EEFF module to mix multi-scale input tokens, returning their concatenated single output, and (c)\,the extended ViDT+ equips the EEFF module for multi-scale feature fusion and the UQR module\,\cite{dong2021solq} for end-to-end multi-task learning. 
}}
\label{fig:vidt_plus_pipeline}
\vspace*{-0.35cm}
\end{figure*}

\smallskip\smallskip
\noindent\textbf{Auxiliary Techniques for Additional Improvements.} The decoder of \algname{} follows the standard structure of multi-layer transformers, generating refined $[\mathtt{DET}]$ tokens at each layer. Hence, \algname{} leverages the two auxiliary techniques used in (Deformable) DETR for additional improvements:
\begin{itemize}[leftmargin=10pt]
\item {Auxiliary Decoding Loss\,\cite{carion2020end}}: Detection heads consisting of two feedforward networks\,(FFNs) for box regression and classification are attached to every decoding layer. All the training losses from detection heads at different scales are added. 
This helps the model output the correct number of objects without non-maximum suppression.
\vspace*{0.1cm}
\item {Iterative Box Refinement\,\cite{zhu2021deformable}}: Each decoding layer refines bounding boxes based on predictions from the detection head in the previous layer. 
Thus, the box regression process progressively improves through the decoding layers.
\end{itemize}

These two techniques are essential for transformer-based object detectors because they significantly enhance detection performance without compromising detection efficiency. We provide an ablation study of their effectiveness for object detection in Section \ref{sec:auxiliary_ablation}.

{\color{black}
\subsection{Complexity Analysis: ViDT vs. YOLOS}
We analyze the computational complexity of the proposed RAM compared with the attention used in YOLOS, based on building blocks of the original and Swin Transformer models \cite{vaswani2017attention,liu2021swin}.

Let {\sf P} and {\sf D} be the number of $[\mathtt{PATCH}]$ and $[\mathtt{DET}]$ tokens (${\sf D} \ll {\sf P}$ in practice, \emph{e.g.}, ${\sf P}=66,650$ and ${\sf D}=100$ at the first stage of ViDT with the resolution of $800\times1333$ images), respectively. The computational complexity of the attention module for YOLOS and ViDT (RAM) is then derived as below, also summarized in Table \ref{table:full_analysis}:

\begin{itemize}[leftmargin=10pt]
\item\textbf{YOLOS Attention}: $[\mathtt{DET}]$ tokens are simply appended to $[\mathtt{PATCH}]$ tokens to perform global self-attention on $[\mathtt{PATCH}, \mathtt{DET}]$ tokens (\emph{i.e.}, ${\sf P} + {\sf D}$ tokens). Thus, the computational complexity is $\mathcal{O}(d^2 ({\sf P} + {\sf D}) + d ({\sf P} + {\sf D})^{2})$, which is \emph{quadratic} to the number of $[\mathtt{PATCH}]$ tokens. If breaking down the total complexity, we obtain $\mathcal{O}\big((d^2 {\sf P} + d {\sf P}^2) +  (d^2 {\sf D} + d{\sf D}^{2}) + d{\sf P}{\sf D}\big)$, where the first and second terms are for the global self-attention for $[\mathtt{PATCH}]$ and $[\mathtt{DET}]$ tokens, respectively, and the last term is for the global cross-attention between them.
\vspace*{0.1cm}
\item\textbf{ViDT (RAM) Attention}: RAM performs the three different attention operations (with the complexities): (1) $[\mathtt{PATCH}]\times[\mathtt{PATCH}]$ local self-attention with window partition, $\mathcal{O}(d^2 {\sf P} + d k^2 {\sf P})$; (2) $[\mathtt{DET}]\times[\mathtt{DET}]$ global self-attention, $\mathcal{O}(d^2 {\sf D} + d {\sf D}^{2})$; (3) $[\mathtt{DET}]\times[\mathtt{PATCH}]$ global cross-attention, $\mathcal{O}(d^2 ({\sf P} + {\sf D}) + d {\sf P} {\sf D})$. In total, the computational complexity of RAM is $\mathcal{O}(d^2 ({\sf P} + {\sf D}) + dk^2 {\sf P} + d {\sf D}^{2} + d {\sf P} {\sf D})$, which is \emph{linear} to the number of $[\mathtt{PATCH}]$ tokens.
\end{itemize}

Consequently, the computational complexity of ViDT\,(RAM) is much lower than that of the attention module used in YOLOS since ${\sf D} \ll {\sf P}$; RAM achieves the \emph{linear} complexity to the patch tokens, while YOLOS suffers from the quadratic complexity. 
}

{\color{black}
\section{ViDT+: Extension to Multi-task Learning}

The proposed ViDT significantly improves both the computational complexity and the performance of the end-to-end transformer-based detector. 
We additionally analyze two drawbacks of ViDT; (1) The multi-scale $\mathtt{PATCH}$ tokens are \textit{linearly} fused, failing to extract complex complementary information; (2) ViDT does not apply to other tasks, such as instance segmentation. 
Resolving the two issues are not trivial. For the former, we need to develop an efficient way of fusing the features non-linearly without compromising inference speed. For the latter, existing FPN-style networks for DETR family detectors cannot be trained in an end-to-end manner for multi-task learning\,\cite{carion2020end}.
We address the drawbacks by adding three components, namely, Efficient Pyramid Feature Fusion\,(EPFF), Unified Query Representation\,(UQR), and IoU-aware and Token Labeling Losses.
The method that incorporates all {three} components is referred to as ViDT+, as illustreated in Figure \ref{fig:vidt_plus_pipeline}(c).

\subsection{Efficient Pyramid Feature Fusion Module}

All the multi-scale $[\mathtt{PATCH}]$ tokens of ViDT are fed into the encoder-free neck component of ViDT without any processing. 
Then, the $[\mathtt{PATCH}]$ tokens are decoded into object embeddings, \emph{i.e.}, the final $[\mathtt{DET}]$ tokens. 
As in Eq.\,\eqref{eq:msdeform_attn}, the multi-scale deformable attention of the decoder fuses the multi-scale $[\mathtt{PATCH}]$ tokens \emph{linearly} via weighted aggregation, but only a few sampled $[\mathtt{PATCH}]$ tokens ($K$ tokens per scale) are used for computational efficiency. 
We thus introduce a simple but efficient pyramid feature fusion\,(EPFF) module, which fuses all the available multi-scale tokens \emph{non-linearly} using multiple CNN fusion blocks before putting them into the decoder, as illustrated in Figure \ref{fig:vidt_plus_pipeline}(a) and \ref{fig:vidt_plus_pipeline}(b). 
The proposed EPFF extracts complementary information from feature maps at different scales more effectively than the previous simple linear aggregation.

Specifically, all the $[\mathtt{PATCH}]$ tokens with multiple scales from the body's different stages are assembled to form multi-scale feature maps $\{{\bm x}^{l}\}_{l=1}^{L}$ with the same size of channel dimension\footnote{{\color{black}We use 256 channel dimension for compatibility with a typical deformable transformer decoder\,\cite{zhu2021deformable}}.}. Subsequently, they are fused in a top-down manner.
Each fusion block in Figure \ref{fig:vidt_plus_pipeline}(a) receives two input feature maps for pyramid feature fusion: (1) a feature map ${\bm x}^{l}$ for a target $l$-th scale (higher resolution) and (2) the fused feature map from the predecessor fusion block for the feature map ${\bm x}^{l+1}$ (lower resolution),
\begin{equation}
{\bm x_{fuse}^{l}} = \overbrace{{\rm ResUnit}}^{{\rm smoothing}} \big( \overbrace{{\rm Upsample}({\bm x}_{fuse}^{l+1})}^{{\rm interpolation}} + \overbrace{{\rm ResUnit}({\bm x}^{l})}^{{\rm transform}}\big),
\end{equation}
%
where the ${\rm Upsample}$ operator resizes the low-resolution feature map to fuse with the high-resolution one via bilinear interpolation, and the two ${\rm ResUnits}$ are the bottleneck residual block for feature transform and feature smoothing, respectively. 
As a result, fused multi-scale features are obtained, flattened along the spatial dimension, and concatenated for all scales as input to the neck decoder, as shown in Figure \ref{fig:vidt_plus_pipeline}(b). 
As analyzed in Section \ref{sec:exp_coco}, this module only adds 1M parameters and greatly increases detection and segmentation accuracy without compromising inference speed.

\subsection{Unified Query Representation Module}

Object detection facilitates joint supervision of multi-task learning and instance segmentation\,\cite{he2017mask, hariharan2014simultaneous, girshick2015region}. 
Hence, we add a unified query representation (UQR) module \cite{dong2021solq} at the beginning of prediction heads, as shown on the right side of Figure \ref{fig:vidt_plus_pipeline}(c).

Unfortunately, $[\mathtt{DET}]$ tokens of the DETR-based approaches only correspond to objects to detect.
Consequently, $[\mathtt{DET}]$ tokens cannot be converted to 2D segmentation tasks, and it makes DETR-based approaches fail to perform object detection and instance segmentation tasks simultaneously\,\cite{dong2021solq}.
To address this issue, Dong et al. \cite{dong2021solq} propose the UQR module, which transforms the ground-truth 2D binary mask of each object into the frequency domain using discrete cosine transform\footnote{{\color{black} Sparse coding and PCA can also be used for the conversion, but DCT is reported to show the best instance segmentation results\,\cite{dong2021solq}.}} (DCT) \cite{ahmed1974discrete}, generating a ground-truth mask \emph{vector} per object to predict. 
Given a ground-truth segmentation mask ${\bm S}$, a ground-truth mask vector ${\bm v}$ is encoded by sampling the low-frequency components from ${\bm F}={\bm A}{\bm S}{\bm A}^{\top}$, where ${\bm A}$ is the transform matrix. 
Hence, the final $[\mathtt{DET}]$ tokens are used directly to predict the ground-truth mask vector using a FFN, which outputs the predicted mask vectors $\hat{{\bm v}}$. 
This procedure is conducted in parallel with classification and box regression for multi-task learning.
As a result, the overall loss function for joint supervision can be formulated as $\ell_{joint} = \ell_{det} + \lambda_{seg}~ \ell_{l1}({\bm v}, \hat{{\bm v}})$, where $\lambda_{seg}$ is a coefficient for the instance segmentation task (see Appendix \ref{appendix:detailed_loss} for details). 
In addition, at evaluation and testing time, the predicted mask vector $\hat{{\bm v}}$ can be converted to the estimated 2D binary mask $\hat{{\bm S}}$ through the inverse sampling and transformation $\hat{{\bm S}}={\bm A}^{-1}\hat{{\bm F}}({\bm A}^{\top})^{-1}$, where $\hat{{\bm F}}$ is the inverse sampling result from $\hat{{\bm v}}$.

\begin{table*}[t]
\begin{center}
\caption{Summary on the ViT backbone. \enquote{dist.} is the distillation strategy for classification\,\cite{touvron2021training}.} 
\vspace*{-2mm}
\label{table:vit_summary}
\small
\begin{tabular}{L{2.8cm} |L{2.5cm} |X{2.6cm} | X{1.3cm} |X{2.0cm} |X{1.7cm} |X{2.3cm}}\hline
\rowcolor{Gray}
Backbone & Type\,(Size) & Train Data & Epochs & Resolution & Params & \!\!ImageNet Acc.\!\! \\\toprule
\multirow{3}{*}{DeiT} 
& {DeiT-tiny\,(dist.)}\!\!     & {ImageNet-1K} & {300} & {$224\times224$} & {6M} & {74.5} \\
& {DeiT-small\,(dist.)}\!\!\!    & {ImageNet-1K} & {300} & {$224\times224$} & {22M} & {81.2} \\
& DeiT-base\,(dist.)\!\!    & ImageNet-1K & 300  & $384\times224$ & 87M & 85.2\\ \midrule
\multirow{4}{*}{\makecell[l]{Swin Transformer}} 
& Swin-nano     & ImageNet-1K & 300 & $224\times224$ & 6M & 74.9  \\
& Swin-tiny     & ImageNet-1K & 300 & $224\times224$ & 28M & 81.2 \\
& Swin-small    & ImageNet-1K & 300 & $224\times224$ & 50M & 83.2 \\ 
& Swin-base     & ImageNet-22K & 90 & $224\times224$ & 88M & 86.3  \\ \bottomrule
\end{tabular}
\end{center}
\vspace*{-4mm}
\end{table*}

\subsection{IoU-aware and Token Labeling Losses}

For dense prediction tasks, the model can capture more diverse aspects of provided inputs when properly incorporating multiple independent objectives. 
Thus, we introduce two additional objectives for training, namely IoU-aware loss and token labeling loss, eventually leading to a considerable performance gain with our proposed ViDT+ model. 
Note that they do not slow down the model inference speed at test time as they are activated only for training:
\begin{itemize}[leftmargin=10pt]
\item {IoU-aware Loss\,\cite{wu2020iou, lin2021d}}: Predicting the IoU score directly using the final $[\mathtt{DET}]$ token helps increase detection confidence, alleviating the mismatch between expected and ground-truth bounding boxes. 
Hence, we add a new FFN branch to predict the IoU score between the predicted bounding box $\hat{b}_i$ and ground-truth one $b_i$. Then, the IoU-aware loss is formulated as
\begin{equation}
\ell_{aware} = \frac{1}{B} \sum_{i=1}^{B} {\rm BCE}\big({\rm FFN}([\mathtt{DET}]_i), {\rm IoU}(b_i, \hat{b}_i)\big),
\label{eq:iou_aware}
\end{equation}
where $[\mathtt{DET}]_i$ is the final $[\mathtt{DET}]$ token (returning from the neck decoder) corresponding to the $i$-th object; and $B$ and ${\rm BCE}$ are the total number of objects in the input image and binary cross-entropy loss function, respectively.
\vspace*{0.10cm}
\item {Token Labeling Loss\,\cite{jiang2021all, lin2021d}}: Token labeling allows to solve multiple token-level recognition problems by assigning each $[\mathtt{PATCH}]$ token with an individual location-specific supervision generated by a machine annotator. 
Here, we leverage the ground-truth segmentation mask to assign the location-specific class label per $[\mathtt{PATCH}]$ token. 
First, the segmentation mask is interpolated to align with the resolution of the feature map generated from the $l$-th stage of the body (\emph{i.e.}, the resolution of ${\bm x}^{l}$); hence, the interpolated mask ${\bm S}^{l}$ is regarded as token-level soft class labels for the $l$-th feature map. Then, the token labeling loss is formulated as
\begin{equation}
\!\!\ell_{token} = \frac{1}{L}\sum_{l=1}^{L}\frac{1}{{P}^{l}}\sum_{i=1}^{{P}^{l}}{\rm Focal}\Big({\rm FFN}\big([\mathtt{PATCH}]^{l}_i\big), {\bm S}^{l}[i]\Big),
\label{eq:token_label}
\end{equation}
where $[\mathtt{PATCH}]^{l}_i$ is the $i$-th $[\mathtt{PATCH}]$ token in the feature map ${\bm x}^{l}$ from the $l$-th stage of the body, and ${\bm S}^{l}[i]$ returns the token-level soft label corresponding to the $[\mathtt{PATCH}]^{l}_i$ token; and $L$ and ${P}^{l}$ is the number of scales and tokens in the feature map ${\bm x}^{l}$, respectively. {\rm Focal} is the focal loss function \cite{lin2017focal} and {\rm FFN} is the classification layer.
\end{itemize}

These two losses are added with their respective coefficients to the joint learning loss if activated. We detail the complete objective of ViDT+ in Appendix \ref{appendix:detailed_loss}.  

}

%% file: 3-evaluation.tex
\section{Evaluation}
\label{sec:evaluation}

{\color{black}
In this section, we present thorough experimental results with evaluations against the state-of-the-art approaches. 
%
}

%

\smallskip\smallskip
\noindent \textbf{Dataset.} We carry out object detection experiments on the Microsoft COCO 2017 benchmark dataset\,\cite{lin2014microsoft}. All the fully transformer-based object detectors are trained on 118K training images and tested on 5K validation images following the standard setups\,\cite{carion2020end}.

\smallskip\smallskip
\noindent \textbf{Algorithms.} We evaluate \algname{} and \algname{}+ against two fully transformer-based object detection pipelines, namely DETR\,(ViT) and YOLOS. 
Since DETR\,(ViT) follows the general pipeline of (Deformable) DETR by replacing its ResNet backbone with other ViT variants, we use one original ViT and one latest ViT variant, DeiT and Swin Transformer as its backbone without any modification. 
%
As YOLOS is strongly coupled with the original ViT architecture, only DeiT is used for evaluation. 
Table \ref{table:vit_summary} summarizes all the ViT models pre-trained on ImageNet used for evaluation. 
Note that publicly available pre-trained models are used except for Swin-nano. 
In addition, we configure Swin-nano\footnote{
Swin-nano is designed such that the number of channels in the hidden layer is half that of Swin-tiny. Please see Appendix \ref{appendix:swin_nano}.} comparable to DeiT-tiny, which is trained on ImageNet with the identical setting. 
Overall, with respect to the number of parameters, Deit-tiny, -small, and -base are comparable to Swin-nano, -tiny, and -base, respectively. 
More details on the evaluated detectors are presented in Appendix \ref{appendix:pipelines}. 

\smallskip
\noindent \textbf{Implementation Details.} All the algorithms are implemented using PyTorch and executed using four NVIDIA Tesla A100 GPUs. 
We train \algname{} and \algname{}+ using AdamW\,\cite{loshchilov2017decoupled} with the same initial learning rate of $10^{-4}$ for its body, neck and head. 
In contrast, following the (Deformable) DETR setting, DETR\,(ViT) is trained with the initial learning rate of $10^{-5}$ for its pre-trained body\,(ViT backbone) and $10^{-4}$ for its neck and head. YOLOS and \algname{}\,(w.o. Neck) are trained with the same initial learning rate of $5\times10^{-5}$, which is the original setting of YOLOS for the neck-free detector. 
We do not change any hyperparameters used in the transformer encoder and decoder for (Deformable) DETR; thus, the neck decoder of \algname{} also consists of six deformable transformer layers using exactly the same hyperparameters. 
{\color{black}
The number of learnable $[\mathtt{DET}]$ tokens are set to be 100 and 300 for \algname{} and \algname{}+, respectively. The effect of different number of $[\mathtt{DET}]$ tokens are discussed in Section \ref{sec:inc_det}.
}
Auxiliary decoding loss and iterative box refinement are applied to the compared methods if applicable.

Regarding the resolution of input images, we use scale augmentation that resizes them such that the shortest side is at least 480 and at most 800 pixels while the longest is at most 1333\,\cite{wu2019detectron2}. 
More details of the experiment configuration can be found in Appendix \ref{appendix:hyperparams}--\ref{appendix:training_config}. 
All the source code and trained models will be made available to the public at \url{https://github.com/naver-ai/vidt}.

\subsection{Experiments with Microsoft COCO Benchmark}
\label{sec:exp_coco}

\begin{table*}[t]
\begin{center}
\caption{Comparison of \algname{} and \algname{}+ with other compared detectors on COCO2017 val set. Two neck-free detectors, YOLOS and ViDT\,(w.o. Neck) are trained for 150 epochs due to the slow convergence. FPS is measured with batch size 1 of $800\times1333$ resolution on a single Tesla A100 GPU, where the value inside the parentheses is measured with batch size 4 of the same resolution to maximize GPU utilization. }
\vspace*{-2mm}
\label{table:full_exp_coco}
\small
\begin{tabular}{L{2.0cm} |L{1.9cm} |X{1.1cm} |X{1.0cm} X{1.0cm} X{1.0cm} X{1.0cm} X{1.0cm} X{1.0cm} |X{1.0cm} |Y{1.5cm}}\hline
\rowcolor{Gray}
Method & \,\,\,\,\,Backbone & \!\!\!\!Epochs\!\!\!\! & AP$^{box}$ & \!\!AP$_{50}^{box}$\!\! & \!\!AP$_{75}^{box}$\!\! & \!\!AP$_{\text{S}}^{box}$\!\! & \!\!AP$_{\text{M}}^{box}$\!\! & \!\!AP$_{\text{L}}^{box}$\!\! & \!\!\!\!Param.\!\!\!\! & \makecell[l]{\hspace*{0.33cm}FPS} \\\toprule
\multirow{7}{*}{DETR} 
& \,\,{DeiT-tiny}     & {50} & {30.0} & {49.2} & {30.5} & {9.9}  & {30.8} & {50.6} & 23M & \!\!{24.4 (28.5)}\!\! \\
& \,\,{DeiT-small}\!\!    & {50} & {32.4} & {52.5} & {33.2} & {11.3} & {33.5} & {53.7} & 39M & \!\!{17.8 (20.3)}\!\! \\ 
& \,\,{DeiT-base}    & {50}& {37.1} & {59.2} & {38.4} & {14.7} & {39.4} & {52.9} & 104M & \!\!{11.1 (14.2)}\!\!\\ 
& \,\,Swin-nano     & 50 & 27.8 & 47.5 & 27.4 & 9.0  & 29.2 & 44.9 & 24M & \!\!50.6 (86.1)\!\! \\
& \,\,Swin-tiny     & 50 & 34.1 & 55.1 & 35.3 & 12.7 & 35.9 & 54.2 & 45M & \!\!39.9 (58.6)\!\!\\
& \,\,Swin-small\!\!    & 50 & 37.6 & 59.0 & 39.0 & 15.9 & 40.1 & 58.9 & 66M & \!\!28.7 (39.2)\!\!\\ 
& \,\,{Swin-base}    & {50} & {40.7} & {62.9} & {42.7} & {18.3} & {44.1} & {62.4} & 104M & \!\!{ 23.4 (30.9)}\!\!\\ \midrule
\multirow{7}{*}{\makecell[l]{Deformable\\DETR}} 
& \,\,{DeiT-tiny}     & {50} & {40.8} & {60.1} & {43.6} & {21.4} & {43.4} & {58.2} & 18M & \!\!{25.1 (36.4)}\!\!\\
& \,\,{DeiT-small}\!\!    & {50} & {43.6} & {63.7} & {46.5} & {23.3} & {47.1} & {62.1} & 34M & \!\!{18.0 (24.0)}\!\!\\ 
& \,\,{DeiT-base}    & {50} & {46.4} & {67.3} & {49.4} & {26.7} & {50.1} & {65.4} & 99M & \!\!{11.3 (13.8)}\!\!\\ 
& \,\,Swin-nano     & 50 & 43.1 & 61.4 & 46.3 & 25.9 & 45.2 & 59.4 & 18M & \!\!14.6 (17.2)\!\!\\
& \,\,Swin-tiny     & 50 & 47.0 & 66.8 & 50.8 & 28.1 & 49.8 & 63.9 & 39M & \!\!13.4 (15.7)\!\!\\
& \,\,Swin-small\!\!    & 50 & 49.0 & 68.9 & 52.9 & 30.3 & 52.8 & 66.6 & 60M & \!\!11.9 (13.9)\!\!\\ 
& \,\,{Swin-base}    & {50} & {51.4} & {71.7} & {56.2} & {34.5} & {55.1} & {67.5} & 98M & \!\!{10.9 (12.7)}\!\! \\ \midrule
\multirow{3}{*}{YOLOS} 
& \,\,DeiT-tiny     & 150 & 30.4 & 48.6 & 31.1 & 12.4 & 31.8 & 48.2 & 6M & \!\!52.5 (61.3)\!\!\\
& \,\,DeiT-small\!\!    & 150 & 36.1 & 55.7 & 37.6 & 15.6 & 38.4 & 55.3 & 30M & \!\!14.0 (24.8)\!\!\\
& \,\,DeiT-base     & 150 & 42.0 & 62.2 & 44.5 & 19.5 & 45.3 & 62.1 & 104M & \!\!~7.0 (12.2)\!\!\\ \midrule
\multirow{4}{*}{\makecell[l]{ViDT\\(w.o. Neck)}} 
& \,\,Swin-nano     & 150 & 28.7 & 48.6 & 28.5 & 12.3 & 30.7 & 44.1 & 7M  & \!\!72.4 (96.9)\!\!\\
& \,\,Swin-tiny     & 150 & 36.3 & 56.3 & 37.8 & 16.4 & 39.0 & 54.3 & 29M & \!\!51.8 (60.4)\!\!\\
& \,\,Swin-small\!\!    & 150 & 41.6 & 62.7 & 43.9 & 20.1 & 45.4 & 59.8 & 52M & \!\!33.5 (38.4)\!\!\\ 
& \,\,Swin-base     & 150 & 43.2 & 64.2 & 45.9 & 21.9 & 46.9 & 63.2 & 91M & \!\!26.3 (29.6)\!\!\\ \midrule
\multirow{4}{*}{{ViDT}} 
& \,\,Swin-nano     & 50 & 40.4 & 59.6 & 43.3 & 23.2 & 42.5 & 55.8 & 15M & \!\!40.8 (76.0)\!\!\\
& \,\,Swin-tiny     & 50 & 44.8 & 64.5 & 48.7 & 25.9 & 47.6 & 62.1 & 37M & \!\!33.5 (51.2)\!\!\\
& \,\,Swin-small\!\!    & 50 & 47.5 & 67.7 & 51.4 & 29.2 & 50.7 & 64.8 & 60M & \!\!24.7 (34.6)\!\!\\ 
& \,\,Swin-base     & 50 & {49.2} & {69.4} & {53.1} & {30.6} & 52.6 & {66.9} & 99M & \!\!20.5 (27.1)\!\!\\ \midrule
\multirow{4}{*}{{{ViDT+}}} 
& \,\,Swin-nano     & 50 & 45.3 & 62.3 & 48.9 & 27.3 & 48.2 & 61.5 & 16M & \!\!37.6 (69.3)\!\!\\
& \,\,Swin-tiny     & 50 & 49.7 & 67.7 & 54.2 & 31.6 & 53.4 & 65.9 & 38M & \!\!30.4 (47.6)\!\!\\
& \,\,Swin-small\!\!    & 50 & 51.2 & 69.5 & 55.9 & 33.8 & 54.5 & 67.8 & 61M & \!\!22.6 (32.4)\!\!\\ 
& \,\,Swin-base     & 50 & {\bf 53.2} & {\bf 71.6} & {\bf 58.3} & {\bf 36.0} & {\bf 57.1} & {\bf 69.2} & 100M & \!\!19.3 (25.7)\!\!\\\bottomrule 
\end{tabular}
\end{center}
\vspace*{-4mm}
\end{table*}

\begin{table*}[t]
\begin{center}
\caption{{Evaluations of \algname{} with DETR and Deformable DETR using the CNN backbone (ResNet-50) on COCO2017 val set.}}
\vspace*{-2mm}
\label{table:resnet_comp}
\small
\begin{tabular}{L{2.5cm} |L{1.5cm} |X{1.0cm} |X{1.0cm} X{1.0cm} X{1.0cm} X{1.0cm} X{1.0cm} X{1.0cm} |X{1.0cm} |Y{1.5cm}}\hline
\rowcolor{Gray}
Method & Backbone & \!\!\!\!Epochs\!\!\!\! & AP$^{box}$ & \!\!AP$_{50}^{box}$\!\! & \!\!AP$_{75}^{box}$\!\! & \!\!AP$_{\text{S}}^{box}$\!\! & \!\!\!AP$_{\text{M}}^{box}$\!\! & \!\!AP$_{\text{L}}^{box}$\!\! & \!\!\!\!Param.\!\!\!\! & \makecell[l]{\hspace*{0.33cm}FPS} \\\toprule
\!{DETR} & {ResNet-50}   & {500} & {42.0} & {62.4} & {44.2} & {20.5} & {45.8} & {61.1} & {41M} & {\!\!54.1 (99.3)}\!\!\\
\!{DETR-DC5} & {ResNet-50}     & {500} & {43.3} & {63.1} & {45.9} & {22.5} & {47.3} & {61.1} & {41M} & \!\!{29.7 (34.3)}\!\!\\
\!{DETR-DC5} & {ResNet-50}    & {50} & {35.3} & {55.7} & {36.8} & {15.2} & {37.5} & {53.6} & {41M} & \!\!{29.7 (34.3)}\!\!\\ 
{\!Deformable DETR\!\!} & {ResNet-50}    & {50} & {45.4} & {64.7} &  {49.0} & {26.8} & {48.3} & {61.7} & {40M} & \!\!{27.2 (43.6)}\!\!\\ \midrule
\!{ViDT} & {Swin-tiny}     & {50} & {44.8} & {64.5} & {48.7} & {25.9} & {47.6} & {62.1} & {37M} & \!\!{33.5 (51.2)}\!\!\\
\!{ViDT} & {Swin-tiny}     & {150} & {47.2} & {66.7} & {51.4} & {28.4} & {50.2} & {64.7} & {37M} & \!\!{33.5 (51.2)}\!\!\\ \bottomrule
\end{tabular}
\end{center}
\vspace*{-4mm}
\end{table*}

\subsubsection{Object Detection}

Table \ref{table:full_exp_coco} shows evaluation results in terms of AP$^{box}$, FPS, and number of parameters, where two variants of DETR\,(ViT) are simply named DETR and Deformable DETR, respectively. 
A summary plot is also provided in Figure \ref{fig:ap_vs_latency}.

\smallskip\smallskip
\noindent \textbf{Highlights.} Among detection-only methods, \algname{} achieves the best trade-off between AP$^{box}$ and FPS. 
With its high scalability, it performs well even with Swin-base of $0.1$ billion parameters, which is 2x faster than Deformable DETR with similar AP$^{box}$.
In addition, \algname{} shows $40.4$AP$^{box}$ only with 15M parameters; it is $6.3$--$12.6$AP$^{box}$ higher than those of DETR\,(swin-nano) and DETR\,(swin-tiny), which exhibit similar FPS of $39.9$--$50.6$.
{\color{black}
ViDT+ shows considerable AP$^{box}$ improvements over the vanilla ViDT with a little decrease in FPS. 
For example, ViDT+\,(swin-nano) achieves 45.3AP$^{box}$ only with 16M parameters, which is even higher than that of ViDT\,(swin-tiny) with 37M parameters. 
In particular, to the best of our knowledge, 53.2AP$^{box}$ of VIDT+\,(Swin-base) is the highest among existing fully transformer-based detectors. 
In the detailed analysis below, we only compare the vanilla ViDT with other detection-only architectures since ViDT+ requires multiple extended modules, such as EPFF and UQR, and leverages extra knowledge from instance segmentation.
}

\begin{table*}[t]
\begin{center}
\caption{Comparison of \algname{}+ with other multi-task learning methods for instance segmentation on COCO2017 val set. To distinguish AP for detection and segmentation, we put `seg' and `box' superscripts into AP at the first row, respectively. Except \algname{}+, all the results are borrowed from \cite{dong2021solq}.  }
\vspace*{-2mm}
\label{table:full_exp_seg}
\small
\begin{tabular}{L{2.45cm} |L{2.45Cm} |X{1.0cm} |X{0.95cm} X{0.95cm} X{0.95cm} X{0.95cm} |X{0.95cm} X{0.95cm} X{0.95cm} X{0.95cm}}\hline
\rowcolor{Gray}
Method & \,\,\,\,\,\,\,\,\,Backbone & \!\!\!\!Epochs\!\!\!\! & AP$^{seg}$ & \!\!AP$_{S}^{seg}$\!\! & \!\!AP$_{M}^{seg}$\!\! & \!\!AP$_{\text{L}}^{seg}$\!\! &  AP$^{box}$ & \!\!AP$_{S}^{box}$\!\! & \!\!AP$_{M}^{box}$\!\! & \!\!AP$_{L}^{box}$\!\! \\\toprule
\!Mask R-CNN \cite{he2017mask}\!\!\!\!   & {ResNet-50+FPN}    & {36} & {37.5} & {21.1} & {39.6} & {48.3} & {41.3} & {24.2} & 43.6 & 51.7\\ 
\!HTC \cite{chen2019hybrid}      & {ResNet-50+FPN}    & {36} & {39.7} & {22.6} & {42.2} & {50.6} & {44.9} & - & - & - \\ 
\!SOLOv2 \cite{wang2020solov2}      & {ResNet-50+FPN}    & {72} & {38.8} & {16.5} & {41.7} & {56.2} & {40.4} & {20.5} & 44.2 & 53.9\\ 
\!QueryInst \cite{fang2021instances}   & {ResNet-50+FPN}    & {36} & {40.6} & {\textbf{23.4}} & {42.5} & {52.8} & {45.6} & - & - & -\\ 
\!SOLQ \cite{dong2021solq}      & {ResNet50}        & {50} & {\textbf{39.7}} & {21.5} & {42.5} & {53.1} & {47.8} & {27.6} & 50.9 & 61.6\\ \midrule
\!ViDT+       & {Swin-tiny}       & {50} & {39.5} & {21.5} & \textbf{43.4} & \textbf{58.2} & \textbf{49.7} & \textbf{31.6} & \textbf{53.4} & \textbf{65.9}\\ \bottomrule\toprule
\!Mask R-CNN \cite{he2017mask}\!\!\!\!  & {ResNet-101+FPN\!\!}   & {36} & {38.8} & {21.8} & {41.4} & {50.5} & {41.3} & {24.2} & 43.6 & 51.7\\ 
\!HTC \cite{chen2019hybrid}        & {ResNet-101+FPN\!\!}   & {36} & {40.8} & {23.0} & {43.5} & {52.6} & {44.3} & - & - & -\\ 
\!SOLOv2 \cite{wang2020solov2}     & {ResNet-101+FPN\!\!}   & {72} & {39.7} & {17.3} & {42.9} & {57.4} & {42.6} & {22.3} & 46.7 & 56.3\\ 
\!QueryInst \cite{fang2021instances}  & {ResNet-101+FPN\!\!}   & {36} & {\textbf{42.8}} & \textbf{24.6} & \textbf{45.0} & {55.5} & {48.1} & - & - & -\\ 
\!SOLQ \cite{dong2021solq}        & {ResNet-101}       & {50} & {40.9} & {22.5} & {43.8} & {54.6} & {48.7} & {28.6} & 51.7 & 63.1\\ \midrule
\!ViDT+       & {Swin-small}      & {50} & {40.8} & {22.6} & 44.3 & \textbf{60.1} & \textbf{51.2} & \textbf{33.8} & \textbf{54.5} & \textbf{67.8}\\ \bottomrule
\end{tabular}
\end{center}
\vspace*{-4mm}
\end{table*}

\begin{table*}[h]
\begin{center}
\caption{AP and FPS comparison with different selective cross-attention strategies.}
\label{table:ablation_cross}
\vspace*{-2mm}
\small
\begin{tabular}{L{2.4cm} |X{1.1cm} X{1.1cm} |X{1.1cm} X{1.1cm} |X{1.1cm} X{1.1cm} |X{1.1cm} X{1.1cm} |X{1.1cm} X{1.1cm}}\hline
\rowcolor{Gray}
Stage Ids & \multicolumn{2}{c|}{\{1,~2,~3,~4\}} & \multicolumn{2}{c|}{\{2,~3,~4\}} & \multicolumn{2}{c|}{\{3,~4\}} & \multicolumn{2}{c|}{\{4\}}  & \multicolumn{2}{c}{\{~\}} \\\toprule
Metric      & AP$^{box}$ & FPS & AP$^{box}$ & FPS & AP$^{box}$ & FPS & \cellcolor{blue!10}AP$^{box}$ & \cellcolor{blue!10}FPS & AP$^{box}$ & FPS \\\midrule
\algname{} w.o. Neck   & 29.0 & 46.9 & 28.8 & 55.2 & 28.5 & 59.5 & \cellcolor{blue!10}28.7 & \cellcolor{blue!10}72.4 & FAIL & 75.8 \\
\algname{} w. \hspace*{0.17cm} Neck & 40.3 & 31.6 & 40.1 & 34.2 & 40.3 & 36.1 & \cellcolor{blue!10}40.4 & \cellcolor{blue!10}40.8 & 37.1 & 41.2 \\\bottomrule
\end{tabular}
\end{center}
\vspace*{-4mm}
\end{table*}

\smallskip\smallskip
\noindent \textbf{ViDT vs. Deformable DETR.} Thanks to the use of multi-scale features, Deformable DETR exhibits high detection performance in general. 
Nevertheless, its encoder and decoder structure in the neck becomes a critical bottleneck in computation. 
In particular, the encoder with multi-layer deformable transformers adds considerable overhead to transform multi-scale features by attention.
Thus, it achieves low FPS although it achieves higher AP$^{box}$ with a relatively small number of parameters. 
In contrast, \algname{} removes the need for a transformer encoder in the neck by using the Swin Transformer with RAM as its body, directly extracting multi-scale features suitable for object detection. \looseness=-1

\smallskip\smallskip
\noindent \textbf{ViDT\,(w.o. Neck) vs. YOLOS.} For the comparison with YOLOS, we train \algname{} without using its neck component. 
These two neck-free detectors achieve relatively low AP$^{box}$ compared with other detectors in general. %
In terms of speed, YOLOS exhibits much lower FPS than ViDT\,(w.o. Neck) because of its quadratic computational complexity for attention. 
However, as ViDT\,(w.o. Neck) extends the Swin Transformers with RAM, it requires linear complexity for attention. 
Hence, it achieves comparable AP$^{box}$ as YOLOS for various backbone sizes, but with higher FPS.

\smallskip\smallskip
\noindent \textbf{Comparison with Other Possible Integration.} It is of great interest to see whether better integration can also be achieved by (1)\,Deformable DETR without its neck encoder {as its neck decoder also has $[\mathtt{DET}]\times[\mathtt{PATCH}]$ cross-attention}, or (2)\,YOLOS with VIDT's neck decoder {because of the use of multiple auxiliary techniques}. 
Such integration is actually not effective; the former significantly drops AP$^{box}$, while the latter has a much greater drop in FPS than an increase in AP$^{box}$. 
The detailed analysis can be found in Appendix \ref{appendix:extra_exp}.

\begin{table}[t!]
\vspace*{-2mm}
\caption{Results for different spatial encodings for $[\mathtt{DET}]\times[\mathtt{PATCH}]$ cross-attention.}
\vspace*{-3mm}
\label{table:ablation_pos}
\small
\begin{tabular}{L{1.3cm} |X{1.0cm} |X{1.0cm} |X{1.0cm} | X{1.0cm} |X{1.0cm}}\hline
\rowcolor{Gray}
Method  & \!\!None\!\! & \multicolumn{2}{c|}{Pre-addition} &  \multicolumn{2}{c}{Post-addition}\\\toprule
Type    &  \!\!None\!\! &\!\!\!\!\!\cellcolor{blue!10}Sin.\!\!\!\!\! & \!\!Learn.\!\!\!\!\! & \!\!\!\!\!Sin.\!\!\!\!\! & \!\!Learn.\!\!\!\!\! \\\midrule
AP$^{box}$      &  \!\!23.7\!\! &\cellcolor{blue!10}\!\!28.7\!\! & 27.4\!\! & \!\!28.0\!\! & \,24.1\!\!\!\! \\\bottomrule
\end{tabular}
\vspace*{-2mm}
\end{table}

{\color{black}
\smallskip\smallskip
\noindent \textbf{Comparison with Detectors using CNN Backbone.} We compare \algname{} with (Deformable) DETR using the ResNet-50 backbone, as summarized in Table \ref{table:resnet_comp}, where all the results except ViDT are borrowed from \cite{carion2020end,zhu2021deformable}, and DETR-DC5 is a modification of DETR to use a dilated convolution at the last stage in ResNet. 
For fair comparisons, we use ViDT (Swin-tiny) with similar parameter numbers in the following experiments. 
In general, \algname{} shows better trade-offs between AP$^{box}$ and FPS even compared with (Deformable) DETR with the ResNet-50. 
Specifically, \algname{} achieves considerably higher FPS than those of DETR-DC5 and Deformable DETR with competitive AP$^{box}$.  
When training ViDT for 150 epochs, \algname{} outperforms other compared methods using the ResNet-50 backbone in terms of both AP$^{box}$ and FPS. 
}

\begin{table}
\small
\vspace*{-2mm}
\begin{center}
    \caption{Effect of auxiliary decoding loss and iterative box refinement loss with YOLOS\,(DeiT-tiny) and ViDT\,(Swin-nano).}
\vspace*{-3mm}
\label{table:ablation_auxiliary}
\begin{tabular}{L{1.2cm} |X{1.0cm} X{1.0cm} X{1.0cm} | X{1.1cm} |L{1.0cm}}\hline
\rowcolor{Gray}
 & \!\!\!\!Aux. $\ell$\!\!\!\! & \!\!\!\!\!\!Box Ref.\!\!\!\!\!\! & \!\!Neck\!\!  &  AP$^{box}$ & \,\,\,\,\,\,$\Delta$ \\\toprule
\multirow{3}{*}{{YOLOS}} 
&               &               &              &  \!\!30.4\!\! &           \\
&\checkmark     &               &              &  \!\!29.2\!\! & \,$-$1.2\!\!        \\
&\checkmark     & \checkmark    &              &  \!\!20.1\!\! & \,$-$10.3\!\!       \\\midrule
\multirow{5}{*}{{ViDT}} 
&               &               &              &  \!\!28.7\!\! &          \\
& \checkmark    &               &              &  \!\!27.2\!\! & \,$-$1.6\!\!       \\
& \checkmark    & \checkmark    &              &  \!\!22.9\!\! & \,$-$5.9\!\!       \\
& \checkmark    &               & \checkmark   &  \!\!36.2\!\! & \,$+$7.4\!\!           \\
& \cellcolor{blue!10}\checkmark    & \cellcolor{blue!10}\checkmark    & \cellcolor{blue!10}\checkmark   &  \cellcolor{blue!10}\!40.4\!\! & \cellcolor{blue!10}\,\,$+$11.6\!\!           \\\bottomrule
\end{tabular}  
\end{center}
\vspace*{-4mm}
\end{table}

\begin{table*}[h]
\begin{center}
\caption{Effect of additional losses and modules for \algname{}+ with Swin-nano. }
\vspace*{-3mm}
\label{table:plus_analysis}
\small
\begin{tabular}{X{1.5cm} X{1.3cm} X{1.1cm} X{1.1cm} |X{0.7cm} |X{0.84cm} X{0.84cm} X{0.84cm} X{0.84cm} X{0.84cm} X{0.84cm} |X{0.85cm} |X{1.16cm} }\hline
\rowcolor{Gray}
\!\!\!\!IoU-aware\!\!\!\! & \!\!\!\!Token Label\!\!\!\! & \!\!\!EPFF\!\!\! & \!\!\!UQR\!\!\! & \!\!\!Epoch\!\!\!  & \!AP$^{box}$\! &  \!\!\!AP$^{box}_{50}$\!\!\! & \!\!\!AP$^{box}_{75}$\!\!\! &  \!\!\!\!AP$^{box}_S$\!\!\!\! & \!AP$^{box}_M$\!  & \!AP$^{box}_L$\! & \!\!\!Param.\!\!\! &  FPS \\\toprule
 &  &  &  & 50 & 40.4 & 59.6 & 43.3 & 23.2 & 42.5 & 55.8 & 15M & \!\!\!40.8 (76.0)\!\!\! \\
\checkmark &  &  &  & 50 & 41.0 & 59.5 & 44.1 & 22.8 & 43.7 & 56.7 & 15M & \!\!\!40.8 (76.0)\!\!\! \\
\checkmark & \checkmark &  &  & 50 & 41.2 & 59.5 & 44.4 & 23.5 & 44.0 & 57.5 & 15M & \!\!\!40.8 (76.0)\!\!\!\\
\checkmark & \checkmark & \checkmark &  & 50 & 42.5 & 60.9 & 45.3 & 23.5 & 45.3 & 59.0 & 16M & \!\!\!37.6 (69.3)\!\!\! \\
\cellcolor{blue!10}\checkmark & \cellcolor{blue!10}\checkmark & \cellcolor{blue!10}\checkmark & \cellcolor{blue!10}\checkmark & \cellcolor{blue!10}50 & \cellcolor{blue!10}45.3 & \cellcolor{blue!10}62.3 & \cellcolor{blue!10}48.9 & \cellcolor{blue!10}27.3 & \cellcolor{blue!10}48.2 & \cellcolor{blue!10}61.5 & \cellcolor{blue!10}16M & \cellcolor{blue!10}\!\!\!37.6 (69.3)\!\!\! \\\bottomrule
\end{tabular}
\end{center}
\vspace*{-4mm}

\end{table*}

\begin{table}
\begin{center} 
\caption{Performance change with different number of detection tokens regarding AP$^{box}$, Param, and FPS.}
\label{table:num_det_token}
\vspace*{-2mm}
\small
\begin{tabular}{L{1.1cm} |X{0.8cm} X{0.8cm} X{0.8cm} | X{0.8cm} X{0.8cm} X{0.8cm}}\hline
\rowcolor{Gray}
\!Model\!\! & \multicolumn{3}{c|}{ViDT (Swin-nano)} &  \multicolumn{3}{c}{ViDT (Swin-tiny)}\\
\rowcolor{Gray}
\!Metric\!\! & \!\!\!AP$^{box}$\!\!\! & \!\!\!\!\!Param.\!\!\!\!\! & \!\!FPS\!\!  & \!\!\!AP$^{box}$\!\!\! & \!\!\!\!\!Param.\!\!\!\!\!  & \!\!FPS\!\! \\\toprule
\!\!$100$ {\scriptsize Tokens}\!\!\! & \!\!\!40.4\!\!\! & \!\!\!15M\!\!\! & \!\!\!40.8\!\!\! & \!\!\!44.8\!\!\! & \!\!\!37M\!\!\! & \!\!\!33.5\!\!\! \\
\cellcolor{blue!10}\!\!$300$ {\scriptsize Tokens}\!\!\! & \cellcolor{blue!10}\!\!\!42.1\!\!\! & \cellcolor{blue!10}\!\!\!15M\!\!\! & \cellcolor{blue!10}\!\!\!40.4\!\!\! & \cellcolor{blue!10}\!\!\!46.7\!\!\! & \cellcolor{blue!10}\!\!\!38M\!\!\! & \cellcolor{blue!10}\!\!\!33.0\!\!\! \\
\!\!$500$ {\scriptsize Tokens}\!\!\! & \!\!\!42.3\!\!\! & \!\!\!16M\!\!\! & \!\!\!40.0\!\!\! & \!\!\!46.9\!\!\! & \!\!\!38M\!\!\! & \!\!\!32.4\!\!\! \\\bottomrule
\end{tabular}
\vspace*{-3mm}
\end{center}
\end{table}

\begin{table}
\caption{Performance change with different number of training epochs regarding AP$^{box}$, Param, and FPS.}
\label{table:num_epochs}
\vspace{-4mm}
\begin{center} 
\small
\begin{tabular}{L{1.1cm} |X{0.8cm} X{0.8cm} X{0.8cm} | X{0.8cm} X{0.8cm} X{0.8cm}}\hline
\rowcolor{Gray}
\!Model\!\! & \multicolumn{3}{c|}{ViDT (Swin-nano)} &  \multicolumn{3}{c}{ViDT (Swin-tiny)}\\
\rowcolor{Gray}
\!Metric\!\! & \!\!\!AP$^{box}$\!\!\! & \!\!\!\!\!Param.\!\!\!\!\! & \!\!FPS\!\!  & \!\!\!AP$^{box}$\!\!\! & \!\!\!\!\!Param.\!\!\!\!\!  & \!\!FPS\!\! \\\toprule
\!\!\!~~50 {\scriptsize Epochs}\!\!\! & \!\!\!40.4\!\!\! & \!\!\!15M\!\!\! & \!\!\!40.8\!\!\! & \!\!\!44.8\!\!\! & \!\!\!37M\!\!\! & \!\!\!33.5\!\!\! \\
\cellcolor{blue!10}\!\!\!150 {\scriptsize Epochs}\!\!\! & \cellcolor{blue!10}\!\!\!42.6\!\!\! & \cellcolor{blue!10}\!\!\!15M\!\!\! & \cellcolor{blue!10}\!\!\!40.8\!\!\! & \cellcolor{blue!10}\!\!\!47.2\!\!\! & \cellcolor{blue!10}\!\!\!37M\!\!\! & \cellcolor{blue!10}\!\!\!33.5\!\!\! \\\bottomrule
\end{tabular}
\vspace*{-4mm}
\end{center}
\end{table}

\subsubsection{Instance Segmentation}

{\color{black}
Table \ref{table:full_exp_seg} shows evaluation results of ViDT+ with other methods for multi-task learning of object detection and instance segmentation, based on AP$^{box}$ and AP$^{seg}$.
For fair comparisons w.r.t the number of parameters, Swin-tiny and Swin-small backbones are used for ViDT+, which have similar numbers of parameters to ResNet-50 and ResNet-101.

For the instance segmentation task, \algname{}+ obtains AP$^{seg}$ comparable to the state-of-the-art methods, such as QueryInst \cite{fang2021instances}, and SOLQ \cite{dong2021solq}. 
It is noteworthy that \algname{}+ achieves the best segmentation performance for medium- and large-size objects, although it is not the best for small-size objects.
This can be attributed to that vector encoding of the 2D segmentation mask using DTC loses detailed information about small objects. 
In addition, \algname{}+ achieves a considerable gain for the object detection task, which can be justified by its  AP$^{box}$ much higher than other multi-task learning methods. 
Quantitatively, the detection performance of \algname{}+ is $1.9$--$9.9$AP$^{box}$ higher than other methods. 
These results show that 
\algname{} can be easily extended to \algname{}+ with better detection performance.
}

\subsection{Ablation Study}
\label{sec:exp_abl}

\subsubsection{Reconfigured Attention Module}
\label{sec:exp_ram}
We extend the Swin Transformer with RAM to extract fine-grained features for object detection without maintaining an additional transformer encoder in the neck. 
We provide an ablation study on the two main considerations for RAM, which lead to high accuracy and speed. 
To reduce the effect of secondary factors, we mainly use our neck-free version, \algname{}\,(w.o. Neck), for the ablation study.

\smallskip\smallskip
\noindent\textbf{Selective $[\mathtt{DET}]\times[\mathtt{PATCH}]$ Cross-Attention.} The addition of cross-attention to the Swin Transformer inevitably entails computational overheads, particularly when the number of $[\mathtt{PATCH}]$ is large. 
To alleviate such overheads, we selectively enable cross-attention in RAM at the last stage of the Swin Transformer; this is shown to greatly improve FPS, but barely drop AP$^{box}$. 
Table \ref{table:ablation_cross} summarizes AP$^{box}$ and FPS when using different selective strategies for the cross-attention, where the Swin Transformer consists of four stages. 
It is interesting that all the strategies exhibit similar AP$^{box}$ as long as cross-attention is activated at the last stage. 
Since features are extracted in a bottom-up manner as they go through the stages, it seems difficult to directly obtain useful information about the target object at the low level of stages.
Thus, only using the last stage is the best design choice in terms of high AP$^{box}$ and FPS due to the smallest number of $[\mathtt{PATCH}]$ tokens. 
Meanwhile, the detection fails completely or the performance significantly drops if all the stages are not involved due to the lack of interaction between $[\mathtt{DET}]$ and $[\mathtt{PATCH}]$ tokens that spatial positional encoding is associated with.
A more detailed analysis of the cross- and self-attention is provided in appendices \ref{appendix:analysis_attention} and \ref{appendix:comp_det}.

\smallskip\smallskip
\noindent\textbf{Spatial Positional Encoding.} Spatial positional encoding is essential for $[\mathtt{DET}]\times[\mathtt{PATCH}]$ attention in RAM. 
Typically, the spatial encoding can be added to the $[\mathtt{PATCH}]$ tokens before or after the projection layer (see Figure \ref{fig:reconfigured_attention}), and 
we call the former \enquote{pre-addition} and the latter \enquote{post-addition}. 
For each one, we can design the encoding in a sinusoidal or learnable manner\,\cite{carion2020end}. 
Table \ref{table:ablation_pos} contrasts the results with different spatial positional encodings with \algname{}\,(w.o. Neck).
Overall, pre-addition results in performance improvement higher than post-addition, and specifically, the sinusoidal encoding is better than the learnable one. 
Thus, the 2D inductive bias of the sinusoidal spatial encoding is more helpful in object detection. In particular,
pre-addition with the sinusoidal encoding increases AP$^{box}$ by $5.0$ compared to not using any encoding.


\begin{table}
\small
\caption{Performance trade-off by decoding layer drop regarding AP$^{box}$, Param, and FPS.}
\label{table:layer_drop}
\vspace{-4mm}
\begin{center} 
\begin{tabular}{L{1.1cm} |X{0.8cm} X{0.8cm} X{0.8cm} | X{0.8cm} X{0.8cm} X{0.8cm}}\hline
\rowcolor{Gray}
\!Model\!\! & \multicolumn{3}{c|}{ViDT (Swin-nano)} &  \multicolumn{3}{c}{ViDT (Swin-tiny)}\\
\rowcolor{Gray}
\!Metric\!\! & \!\!\!AP$^{box}$\!\!\! & \!\!\!\!\!Param.\!\!\!\!\! & \!\!FPS\!\!  & \!\!\!AP$^{box}$\!\!\! & \!\!\!\!\!Param.\!\!\!\!\!  & \!\!FPS\!\! \\\toprule
\!0 Drop\!\! & \!\!\!40.4\!\!\! & \!\!\!15M\!\!\! & \!\!\!40.8\!\!\! & \!\!\!44.8\!\!\! & \!\!\!37M\!\!\! & \!\!\!33.5\!\!\! \\
\!1 Drop\!\! & \!\!\!40.2\!\!\! & \!\!\!14M\!\!\! & \!\!\!43.7\!\!\! & \!\!\!44.8\!\!\! & \!\!\!36M\!\!\! & \!\!\!35.2\!\!\! \\
\cellcolor{blue!10}\!2 Drop\!\! & \cellcolor{blue!10}\!\!\!40.0\!\!\! & \cellcolor{blue!10}\!\!\!13M\!\!\! & \cellcolor{blue!10}\!\!\!46.7\!\!\! & \cellcolor{blue!10}\!\!\!44.5\!\!\! & \cellcolor{blue!10}\!\!\!35M\!\!\! & \cellcolor{blue!10}\!\!\!36.7\!\!\! \\
\!3 Drop\!\! & \!\!\!38.6\!\!\! & \!\!\!12M\!\!\! & \!\!\!48.8\!\!\! & \!\!\!43.6\!\!\! & \!\!\!34M\!\!\! & \!\!\!38.5\!\!\! \\
\!4 Drop\!\! & \!\!\!36.8\!\!\! & \!\!\!11M\!\!\! & \!\!\!55.3\!\!\! & \!\!\!41.9\!\!\! & \!\!\!33M\!\!\! & \!\!\!41.0\!\!\! \\
\!5 Drop\!\! & \!\!\!32.5\!\!\! & \!\!\!9M\!\!\! & \!\!\!57.6\!\!\! & \!\!\!38.0\!\!\! & \!\!\!32M\!\!\! & \!\!\!43.2\!\!\! \\\bottomrule
\end{tabular}
\vspace*{-0.4cm}
\end{center}
\end{table}

\begin{table*}[h]
\begin{center}
\caption{Complete component analysis. ``{\color{brown}Brown}'' and ``{\color{teal}Teal}'' colors indicate the performance of vanilla ViDT and its extension to ViDT+, respectively. ``{\color{violet}Violet}'' color indicates the performance of fully optimized ViDT+.}
\vspace*{-0.3cm}
\label{table:complete_analysis}
\small
\begin{tabular}{L{0.4cm} |L{4.15cm} |X{1.0cm} X{1.0cm} X{1.0cm} |X{1.0cm} X{1.0cm} X{1.0cm} |X{1.0cm} X{1.0cm} X{1.0cm} }\hline
\rowcolor{Gray}
& & \multicolumn{3}{c|}{ViDT+ (Swin-nano)} & \multicolumn{3}{c|}{ViDT+ (Swin-tiny)} & \multicolumn{3}{c}{ViDT+ (Swin-small)} \\
\rowcolor{Gray}~\#\! & \hspace*{0.15cm}Added Module/Technique & AP$^{box}$ & Param. & FPS & AP$^{box}$ & Param. & FPS & AP$^{box}$ & Param. & FPS \\\toprule
(1)\!& + RAM                        & 28.7 & 7M & 72.4 & 36.3 & 29M & 51.8 & 41.6 & 52M & 33.5 \\
(2)\!& + Encoder-free Neck          & \cellcolor{brown!20}40.4 & \cellcolor{brown!20}15M & \cellcolor{brown!20}40.8 & \cellcolor{brown!20}44.8 & \cellcolor{brown!20}37M & \cellcolor{brown!20}33.5 & \cellcolor{brown!20}47.5 & \cellcolor{brown!20}60M & \cellcolor{brown!20}24.7 \\
(3)\!& + IoU-aware \& Token Label   & 41.0 & 15M & 40.8 & 45.9 & 37M & 33.5 & 48.5 & 60M & 24.7 \\
(4)\!& + EPFF Module                & 42.5 & 16M & 38.0 & 47.1 & 38M & 30.9 & 49.3 & 61M & 23.0 \\
(5)\!& + UQR Module                 & 43.9 & 16M & 38.0 & 47.9 & 38M & 30.9 & 50.1 & 61M & 23.0 \\
(6)\!& + $300$ $[\mathtt{DET}]$ Tokens    & \cellcolor{teal!20}45.3 & \cellcolor{teal!20}16M & \cellcolor{teal!20}37.6 & \cellcolor{teal!20}49.7 & \cellcolor{teal!20}38M & \cellcolor{teal!20}30.4 & \cellcolor{teal!20}51.2 & \cellcolor{teal!20}61M & \cellcolor{teal!20}22.6 \\
(7)\!& + $150$ Training Epochs          & 47.6 & 16M & 37.6 & 51.4 & 38M & 30.4 & 52.3 & 61M & 22.6 \\
(8)\!&+ Decoding Layer Drop & \cellcolor{violet!15}47.0 & \cellcolor{violet!15}14M & \cellcolor{violet!15}41.9 & \cellcolor{violet!15}50.8& \cellcolor{violet!15}36M & \cellcolor{violet!15}33.9 & \cellcolor{violet!15}51.8 & \cellcolor{violet!15}59M & \cellcolor{violet!15}24.6\\\bottomrule
\end{tabular}
\end{center}
\vspace*{-4mm}
\end{table*}
\subsubsection{Auxiliary Decoding and Iterative Box Refinement}
\label{sec:auxiliary_ablation}

We analyze the performance improvement of auxiliary decoding loss and iterative box refinement. 
To validate the efficacy of these two modules, we extend them for the neck-free detector such as YOLOS where 
each is applied to the encoding layers in the body, as opposed to the conventional way of using the decoding layers in the neck. 
Table\,\ref{table:ablation_auxiliary} shows the performance of the two neck-free detectors, YOLOS and \algname{}\,(w.o. Neck), decreases considerably with the two techniques.
The use of these two modules in the encoding layers is likely to negatively affect the feature extraction of the transformer encoder. 
In contrast, an opposite trend is observed with the neck component. 
Since the neck decoder is decoupled with the feature extraction in the body, the two techniques make a synergistic effect and thus show significant improvements in AP$^{box}$.
These results demonstrate the use of the neck decoder in \algname{} to improve object detection performance. \looseness=-1

{\color{black}
\subsubsection{Components for extension to ViDT+}
\label{sec:exp_abl_extension}

For the extension to \algname{}+, two additional modules are incorporated on top of the vanilla \algname{}, \emph{i.e.}, EPFF module for pyramid feature fusion and UQR module for multi-task learning, and also two independent objectives are added for better optimization, \emph{i.e.}, IoU-aware and token labeling losses.
Table \ref{table:plus_analysis} summarizes the performance improvement of ViDT\,(Swin-nano) when adding them incrementally to the vanilla ViDT; the trend of performance improvement is almost the same for Swin backbones of difference sizes.
IoU-aware and token labeling losses are added first since they do not affect the inference time. 
Then, the remaining EPFF and UQR modules are added to support effective multi-task learning. 
With the Swin-nano backbone, the extension to ViDT+ only adds 1M parameters but AP$^{box}$ improves from 40.4 to 45.3. 
This is a significant performance gain for the better trade-off between accuracy and speed, considering that the runtime performance dropped by 3.2 FPS.

\subsection{Optimization to Performance Boosting}
\label{sec:extra_boosting}

The number of detection tokens and training epochs affect the detection performance w.r.t AP$^{box}$ and FPS. 
Hence, we analyze the performance change by using different numbers of them. 
In addition, we introduce a decoding layer dropping scheme, which further increases the FPS of the ViDT model by simply dropping a few decoding layers in the neck component without compromising AP$^{box}$.  

\subsubsection{Increasing the Number of Detection Tokens}
\label{sec:inc_det}
 
Table \ref{table:num_det_token} shows the performance change of ViDT with different number of $[\mathtt{DET}]$ tokens. 
When used 200 more $[\mathtt{DET}]$ tokens, \emph{i.e.}, 300 tokens, the detection accuracy of \algname{} improves by $1.7$--$1.9$AP$^{box}$ with little increase in FPS.
Thus, the accuracy gain outweighs the loss of efficiency. 
However, there is only a slight improvement in AP$^{box}$ when using 500 $[\mathtt{DET}]$ rather than 300 $[\mathtt{DET}]$ tokens, although FPS decreases almost linearly. 
%
Therefore, we can use 300 $[\mathtt{DET}]$ tokens for a better trade-off between accuracy and speed.

\subsubsection{Increasing the Number of Training Epochs}
\label{sec:inc_epo}

Table \ref{table:num_epochs} shows the performance of ViDT trained with a different number of epochs. 
The increase in the number of training epochs provides performance improvement in ViDT. 
When used 3 times longer training epochs, \emph{i.e.}, 50 epochs $\rightarrow$ 150 epochs, the detection accuracy of \algname{} improves by $2.2$--$2.4$AP$^{box}$ without any performance drop in FPS. 
When the computational budget allows, \algname{} has the potential to achieve better detection performance.
We report the performance of ViDT+ with longer training epochs in our complete analysis of Section \ref{sec:exp_component}.
}

\subsubsection{Decoding Layer Dropping}
\label{sec:dropping_ablation}

ViDT has six layers of transformers as its neck decoder. 
We emphasize that not all layers of the decoder are required at inference time for high performance. 
Table \ref{table:layer_drop} shows the performance of \algname{} when dropping its decoding layer one by one from the top in the inference step. 
Although there is a trade-off relationship between accuracy and speed as the layers are detached from the model, there is no significant AP$^{box}$ drop even when the two layers are removed. 
Although this scheme is not designed for performance evaluation with other methods in Table \ref{table:full_exp_coco} with other methods, we can accelerate the inference speed of a trained \algname{} model to over $10\%$ by dropping its two decoding layers without a much decrease in AP$^{box}$.

{\color{black}
\subsection{Complete Component Analysis}
\label{sec:exp_component}

In this section, we combine all the proposed components (even with longer training epochs and decoding layer drop) to achieve high accuracy and speed for object detection. 
As summarized in Table \ref{table:complete_analysis}, there are \emph{eight} components for extension: (1)\,RAM to extend Swin Transformer as a standalone object detector, (2) the neck decoder to exploit multi-scale features with two auxiliary techniques, (3) the IoU-aware and token labeling losses for fine-grained supervision per token, (4) the EPFF module to fuse multi-scale features non-linearly via pyramid feature fusion, (5) the UQR model for extra supervision from multi-task learning, (6) the use of more detection tokens, (7) the use of longer training epochs, and (8) decoding layer drop to further accelerate inference speed. 
For the full model, it achieves 47.0AP$^{box}$ with very high FPS by only using $14$M parameters when using Swin-nano as its backbone. 
Further, it achieves  50.8--51.8AP$^{box}$ with reasonable FPS when using Swin-tiny and Swin-small backbones.
This indicates that a fully transformer-based object detector has the potential to be used as a generic object detector when further developed in the future.
}

%% file: 4-conclusion.tex
\section{Conclusion}
\label{sec:conclusion}

In this work, we explore the integration of vision and detection transformers to build an effective and efficient object detector. 
The proposed \algname{} significantly improves the scalability and flexibility of transformer models to achieve high accuracy and inference speed. 
The computational complexity of its attention modules is linear \wrt image size, and \algname{} synergizes several essential techniques to boost the detection performance. 
Further, it can be easily extended to support multi-task learning of object detection and instance segmentation. 
The joint learning framework named \algname{}+ improves its detection performance considerably without compromising its efficiency.
On the Microsoft COCO benchmark, \algname{}+ achieves $53.2$AP$^{box}$ with a large Swin-base backbone, and $47.0$AP$^{box}$ with the smallest Swin-nano backbone using only 14M parameters, suggesting the potential of using the proposed model for complex computer vision tasks. \looseness=-1

%% file: 6-appendix.tex
\appendices

\renewcommand{\thepage}{\arabic{page}} 
\setcounter{page}{1}
\setcounter{figure}{0}
\setcounter{table}{0}
\setcounter{equation}{0}

\section{Experimental Details}
\label{appendix:exp_setting}
\vspace*{0.1cm}

\subsection{Swin-nano Architecture}
\label{appendix:swin_nano}
\vspace*{-0.1cm}

Due to the absence of Swin models comparable to Deit-tiny, we configure Swin-nano, which is a 0.25$\times$ model of Swin-tiny such that it has $6$M training parameters comparable to Deit-tiny. Table \ref{table:swin_architecture} summarizes the configuration of Swin Transformer models available, including the newly introduced Swin-nano; S1--S4 indicates the four stages in Swin Transformer. The performance of all the pre-trained Swin Transformer models are summarized in Table \ref{table:vit_summary} in the manuscript.

\vspace*{-0.1cm}
\subsection{Detection Pipelines of All Compared Detectors}
\label{appendix:pipelines}

All the compared fully transformer-based detectors are composed of either (1)\,\emph{body--neck--head} or (2)\, \emph{body--head} structure, as summarized in Table \ref{table:pipeline_summary}. The main difference of \algname{} is the use of reconfigured attention modules\,(RAM) for Swin Transformer, allowing the extraction of fine-grained detection features directly from the input image. Thus, Swin Transformer is extended to a standalone object detector called \algname{}\,(w.o. Neck). Further, its extension to \algname{} allows to use multi-scale features and multiple essential techniques for better detection, such as auxiliary decoding loss and iterative box refinement, by only maintaining a transformer decoder at the neck. Except for the two neck-free detector, YOLOS and \algname{}\,(w.o. Neck), all the pipelines maintain multiple FFNs; that is, a single FFNs for each decoding layer at the neck for box regression and classification.

We believe that our proposed RAM can be combined with even other latest efficient vision transformer architectures, such as PiT\,\cite{heo2021rethinking}, PVT\,\cite{wang2021pyramid} and Cross-ViT\,\cite{chen2021crossvit}. We leave this as future work.

\subsection{Hyperparameters of Neck Transformers}
\label{appendix:hyperparams}

The transformer decoder at the neck in \algname{} introduces multiple hyperparameters. 
We follow exactly the same setting used in Deformable DETR. 
Specifically, we use six layers of deformable transformers with width 256; thus, the channel dimension of the $[\mathtt{PATCH}]$ and $[\mathtt{DET}]$ tokens extracted from Swin Transformer are reduced to 256 to be utilized as compact inputs to the decoder transformer. For each transformer layer, multi-head attention with eight heads is applied, followed by the point-wise FFNs of 1024 hidden units. Furthermore, an additive dropout of 0.1 is applied before the layer normalization. All the weights in the decoder are initialized with Xavier initialization. For (Deformable) DETR, the tranformer decoder receives a fixed number of learnable detection tokens. We set the number of detection tokens to $100$, which is the same number used for YOLOS and \algname{}.

\subsection{Implementation}
\label{appendix:detailed_implementation}

\subsubsection{Detection Head for Prediction}

The last $[\mathtt{DET}]$ tokens produced by the body or neck are fed to a $3$-layer FFNs for bounding box regression and linear projection for classification,
\begin{equation}
\normalsize
\hat{{{B}}} = \text{FFN}_{\text{3-layer}}\big([\mathtt{DET}]\big) ~~ \text{and} ~~ \hat{{{P}}} = \text{Linear}\big([\mathtt{DET}]\big).
\end{equation}

For box regression, the FFNs produce the bounding box coordinates for $d$ objects, $\hat{B}\in[0,1]^{d \times 4}$, that encodes the normalized box center coordinates along with its width and height. For classification, the linear projection uses a softmax function to produce the classification probabilities for all possible classes including the background class, $\hat{P} \in [0,1]^{d \times (c+1)}$, where $c$ is the number of object classes. When deformable attention is used on the neck in Table \ref{table:pipeline_summary}, only $c$ classes are considered without the background class for classification. This is the original setting used in DETR, YOLOS\,\cite{carion2020end, fang2021you} and Deformable DETR\,\cite{ zhu2021deformable}.

\begin{table}[t]
\caption{Swin Transformer Architecture, where S1, S2, S3, and S4 stand for the number of stages.}
\vspace*{-0.4cm}
\label{table:swin_architecture}
\begin{center}
\small
\vspace*{0.25cm}
\begin{tabular}{L{1.8cm} |X{1.5cm} |X{0.74cm} X{0.74cm} X{0.74cm} X{0.74cm}}\hline
\rowcolor{Gray}
\!Model\!\! & {\!\!\!Channel\!\!\!} & \multicolumn{4}{c}{Stage Numbers}    \\ 
\rowcolor{Gray} 
\!Name\!\! & {\!\!\!Dim.\!\!\!} & \!\!\!\!S1\!\!\!\! & \!\!\!\!S2\!\!\!\! & \!\!\!\!S3\!\!\!\! & \!\!\!\!S4\!\!\!\! \\ \toprule
\!Swin-nano\!\! & 48 & 2 & 2 & 6 & 2 \\
\!Swin-tiny\!\! & 96 & 2 & 2 & 6 & 2 \\
\!Swin-small\!\!& 128 & 2 & 2 & \!18\! & 2 \\
\!Swin-base\!\! & 192 & 2 & 2 & \!18\! & 2 \\\bottomrule
\end{tabular}
\vspace*{-0.3cm}
\end{center}
\end{table}

 \newcommand{\cmark}{$\bigcirc$}%
\newcommand{\xmark}{\ding{53}}%
\begin{table*}
\caption{Comparison of detection pipelines for all available fully transformer-based object detectors, where $\dagger$ indicates that multi-scale deformable attention is used for neck transformers.}
\label{table:pipeline_summary}
\vspace*{-0.4cm}
\begin{center}
\small
\begin{tabular}{L{4.5cm} |L{4.5cm} |X{2.0cm} X{2.0cm} |X{3.0cm}}\hline
\rowcolor{Gray}
\!Pipeline     \!\! & \hspace*{1.9cm}Body  & \multicolumn{2}{c|}{Neck} &  Head  \\
\rowcolor{Gray}
\!Method Name\!\!\!\!   & \hspace*{1.1cm}Feature Extractor   & \!\!\!Tran. Encoder\!\!\!   & \!\!\!Tran. Decoder\!\!\! &  \hspace*{0.4cm}Prediction \\\toprule
\!DETR\,(DeiT)\!\!\!\!  & DeiT Transformer & \cmark & \cmark & \,\,Multiple FFNs\\
\!DETR\,(Swin)\!\!\!\!  & Swin Transformer & \cmark &  \cmark & \,\,Multiple FFNs\\
\!Deformable DETR\,(DeiT)\!\!\!\! & DeiT Transformer & \,\,\cmark$^{\dagger}$ & \,\,\cmark$^{\dagger}$ & \,\,Multiple FFNs \\
\!Deformable DETR\,(Swin)\!\!\!\! & Swin Transformer & \,\,\cmark$^{\dagger}$ & \,\,\cmark$^{\dagger}$ & \,\,Multiple FFNs\\\midrule
\!YOLOS\!\! & DeiT Transformer & \xmark & \xmark & \,\,Single FFNs\\\midrule
\!\algname{}\,(w.o. Neck)\!\! & Swin Transformer$+$RAM\!\! & \xmark & \xmark & \,\,Single FFNs\\
\!\algname{}\!\! & Swin Transformer$+$RAM\!\!  & \xmark & \,\,\cmark$^{\dagger}$ & \,\,Multiple FFNs\\\bottomrule
\end{tabular}
\end{center}
\vspace*{-0.3cm}
\end{table*}

\begin{table*}[t]
\caption{{Variations of Deformable DETR, YOLOS, and \algname{} with respect to their neck structure. They are trained for 50 epochs with the same configuration used in our main experimental results.}}
\vspace*{-0.4cm}
\label{table:exp_variations}
\begin{center}
\small
\begin{tabular}{L{3.4cm} |X{1.8cm} |X{1.0cm} X{1.0cm} X{1.0cm} X{1.0cm} X{1.0cm} X{1.0cm} |X{1.1cm} |Y{1.5cm}}\hline
\rowcolor{Gray}
Method & \!\!\!Backbone\!\! & AP & \!\!AP$_{50}$\!\! & \!\!AP$_{75}$\!\! & \!\!AP$_{\text{S}}$\!\! & \!\!AP$_{\text{M}}$\!\! & \!\!AP$_{\text{L}}$\!\! & \!\!\!\!Param.\!\!\!\! & FPS~~~ \\\toprule
Deformable DETR                  & \makecell[l]{\vspace*{-0.4cm}Swin-nano}   & 43.1 & 61.4 & 46.3 & 25.9 & 45.2 & 59.4 & 17M & 14.6 (17.2)   \\
{$-$ neck encoder}                 &                & {34.0} & {52.8} & {35.6} & {18.0} & {36.3} & {48.4} & {14M} & {42.9 (81.4)} \\\midrule
YOLOS                            & \makecell[l]{\vspace*{-0.4cm}DeiT-tiny}      & 30.4 & 48.6 & 31.1 & 12.4 & 31.8 & 48.2 & 6M  & 52.5 (61.3) \\
$+$ neck decoder                 &                & 38.1 & 57.1 & 40.2 & 20.1 & 40.2 & 56.0 & 13M & 33.6 (52.3)     \\\midrule
\algname{}                       & \makecell[l]{\vspace*{-0.4cm}Swin-nano}      & 40.4 & 59.6 & 43.3 & 23.2 & 42.5 & 55.8 & 15M & 40.8 (76.0) \\
$+$ neck encoder                 &                & 46.1 & 64.1 & 49.7 & 28.5 & 48.7 & 61.7 & 19M & 26.4 (40.3)   \\\bottomrule
\end{tabular}
\end{center}
\vspace*{-0.3cm}
\end{table*}

\begin{table*}[t]
\caption{Comparison of ViDT combined with varying ViT backbones on COCO2017 val set. FPS is measured with batch size 1 of $800\times1333$ resolution on a single Tesla A100 GPU, where the value inside the parentheses is measured with batch size 4 of the same resolution to maximize GPU utilization. }
\vspace*{-0.4cm}
\label{table:vidt_w_other}
\begin{center}
\small
\begin{tabular}{L{2.3cm} |X{1.35cm} |X{1.25cm} X{1.25cm} X{1.25cm} X{1.25cm} X{1.25cm} X{1.25cm} |X{1.2cm} |Y{1.5cm}}\hline
\rowcolor{Gray}
Backbone & \!\!\!\!Epochs\!\!\!\! & AP$^{box}$ & \!\!AP$_{50}^{box}$\!\! & \!\!AP$_{75}^{box}$\!\! & \!\!AP$_{\text{S}}^{box}$\!\! & \!\!AP$_{\text{M}}^{box}$\!\! & \!\!AP$_{\text{L}}^{box}$\!\! & \!\!\!\!Param.\!\!\!\! & \makecell[l]{\hspace*{0.4cm}FPS} \\\toprule
Swin-nano         & 50 & 40.4 & 59.6 & 43.3 & 23.2 & 42.5 & 55.8 & 15M & \!\!40.8 (76.0)\!\!\\
Swin-tiny         & 50 & 44.8 & 64.5 & 48.7 & 25.9 & 47.6 & 62.1 & 37M & \!\!33.5 (51.2)\!\!\\\midrule
CoaT-lite-tiny    & 50 & 41.0 & 59.8 & 44.1 & 23.7 & 43.7 & 56.2 & 13M & \!\!35.0 (72.9)\!\!\\
CoaT-lite-small   & 50 & 44.0 & 63.0 & 47.1 & 26.7 & 46.6 & 60.6 & 28M & \!\!24.8 (46.4)\!\!\\\midrule
PVT-v2-b0         & 50 & 39.7 & 58.3 & 42.4 & 22.4 & 42.0 & 56.0 & 11M & \!\!39.7 (75.0)\!\!\\ 
PVT-v2-b2         & 50 & 47.7 & 67.2 & 51.6 & 28.5 & 50.8 & 64.8 & 35M & \!\!22.4 (35.1)\!\!\\ \bottomrule 
\end{tabular}
\end{center}
\vspace*{-0.3cm}
\end{table*}

\subsubsection{Loss Function for Training}
\label{appendix:detailed_loss}

All the methods adopts the loss function of \,(Deformable) DETR. Since the detection head return a fixed-size set of $d$ bounding boxes, where $d$ is usually larger than the number of actual objects in an image, Hungarian matching is used to find a bipartite matching between the predicted box $\hat{B}$ and the ground-truth box ${B}$. In total, there are three types of training loss: a classification loss $\ell_{cl}$\footnote{Cross-entropy loss is used with standard transformer architectures, while focal loss\,\cite{lin2017focal} is used with deformable transformer architecture.}, a box distance $\ell_{l_{1}}$, and a GIoU loss $\ell_{iou}$\,\cite{rezatofighi2019generalized},
 \begin{equation}
\begin{gathered}
\ell_{cl}(i) = -\text{log}~\hat{{P}}_{\sigma(i),c_i},~~~\ell_{\ell_{1}}(i) =  ||B_{i}-\hat{B}_{\sigma(i)}||_{1},~\text{and}\\
\ell_{iou}(i) = 1 \!-\! \big( \frac{|B_{i} \cap  \hat{B}_{\sigma(i)}|}{|B_{i} \cup  \hat{B}_{\sigma(i)}|} - \frac{|{\sf B}(B_{i}, \hat{B}_{\sigma(i)}) \symbol{92} B_{i} \cup  \hat{B}_{\sigma(i)}| }{|{\sf B}(B_{i}, \hat{B}_{\sigma(i)})|} \big),
\end{gathered}
\end{equation}
where $c_i$ and $\sigma(i)$ are the target class label and bipartite assignment of the $i$-th ground-truth box, and ${\sf B}$ returns the largest box containing two given boxes. Thus, the final loss of object detection is a linear combination of the three types of training loss,
\begin{equation}
\ell_{det} = \lambda_{cl}\ell_{cl} + \lambda_{\ell_{1}}\ell_{l_{1}} +\lambda_{iou} \ell_{iou}.
\label{eq:final_loss}
\end{equation}
The coefficient for each training loss is set to be $\lambda_{cl}=1$, $\lambda_{\ell_{1}}=5$, and $\lambda_{iou}=2$. If we leverage auxiliary decoding loss, the final loss is computed for every detection head separately and merged with equal importance.
%
{\color{black}
Furthermore, if \algname{} is extended to a multi-task learning framework named \algname{}+, the model is trained via the joint-learning loss composed of detection and segmentation losses, introducing three additional coefficients $\lambda_{seg}, \lambda_{aware}$, and $\lambda_{token}$,
\begin{equation}
\begin{gathered}
\ell_{joint} = \ell_{det} + \lambda_{seg} \ell_{l_{seg}}\\ + \lambda_{aware} \ell_{aware} + \lambda_{token} \ell_{token}, 
\end{gathered}
\end{equation}
where $\ell_{seg}$ is the l1 loss between predicted and ground-truth segmentation vectors, $\ell_{aware}$ and $\ell_{token}$ are the IoU-aware and token labeling losses in Eqs. \eqref{eq:iou_aware} and \eqref{eq:token_label}. Their coefficients are set to be $\lambda_{seg}=3.0$, $\lambda_{aware}=2.0$, and $\lambda_{token}=2.0$, respectively.
}

\subsection{Training Configuration}
\label{appendix:training_config}

We train \algname{} for 50 epochs using AdamW\,\cite{loshchilov2017decoupled} with the same initial learning rate of $10^{-4}$ for its body, neck and head. The learning rate is decayed by cosine annealing with batch size of $16$, weight decay of $1\times10^{-4}$, and gradient clipping of $0.1$. In contrast, \algname{}\,(w.o. Neck) is trained for 150 epochs using AdamW with the initial learning rate of $5\times10^{-5}$ by cosine annealing. The remaining configuration is the same as for \algname{}.

Regarding DETR\,(ViT), we follow the setting of Deformable DETR. 
Thus, all the variants of this pipeline are trained for 50 epochs with the initial learning rate of $10^{-5}$ for its pre-trained body\,(ViT backbone) and $10^{-4}$ for its neck and head. Their learning rates are decayed at the 40-th epoch by a factor of 0.1. Meanwhile, the results of YOLOS are borrowed from the original paper\,\cite{fang2021you} except  YOLOS\,(DeiT-tiny); since the result of YOLOS\,(DeiT-tiny) for $800\times1333$ is not reported in the paper, we train it by following the training configuration suggested by authors.

\section{Supplementary Evaluation}

{\color{black}
\subsection{ViDT Combined with Other ViT Backbones}
\label{appendix:ram_other_vit}

The proposed RAM allows for any ViT variants to be a standalone object detector by applying their attention mechanisms to $[\mathtt{PATCH}]\times[\mathtt{PATCH}]$ attention. That is, the Swin's attention is simply replaced with the other one. The rest $[\mathtt{DET}]\times [\mathtt{DET}]$ and $[\mathtt{DET}]\times [\mathtt{PATCH}]$ attention operations are exactly the same regardless of ViT types. By doing this, we combine ViDT with two other state-of-the-art ViT architectures for object detection, namely CoaT \cite{xu2021co}, and PVT-v2 \cite{wang2021pvtv2}. Table \ref{table:vidt_w_other} summarizes their performance compared with ViDT (Swin) on COCO2017 benchmark data. For a fair comparison, we use the backbones whose model size is in the range of `nano' -- `small'.

For the models with 10M--15M parameters, all the backbones show similar AP$^{box}$ of $39.7$--$41.0$, but Swin-nano and PVT-v2-b0 are more efficient than CoaT-lite-tiny in terms of FPS. On the other hand, for the models with 28M--35M parameters, Swin-tiny provides the best trade-off between AP and FPS. Although PVT-v2-b2 achieves the highest AP$^{box}$ of $47.7$, its FPS is $11.1$ lower than Swin-tiny. Coat-lite-small shows FPS similar to PVT-v2-b2 but its AP is very poor compared with other counterparts. Overall, Swin is the most efficient in terms of FPS, while PVT-v2 is the most effective in terms of the number of parameters.
}

\subsection{Variations of Existing Pipelines}
\label{appendix:extra_exp}

We study more variations of existing detection methods by modifying their original pipelines in Table \ref{table:pipeline_summary}. Thus, we remove the neck encoder of Deformable DETR to increase its efficiency, while adding a neck decoder to YOLOS to leverage multi-scale features along with auxiliary decoding loss and iterative box refinement. Note that these modified versions follow exactly the same detection pipeline with \algname{}, maintaining a \emph{encoder-free} neck between their body and head. Table \ref{table:exp_variations} summarizes the performance of all the variations in terms of AP, FPS, and the number of parameters. \looseness=-1

\begin{figure*}[!t]
\begin{center}
\includegraphics[height=110mm]{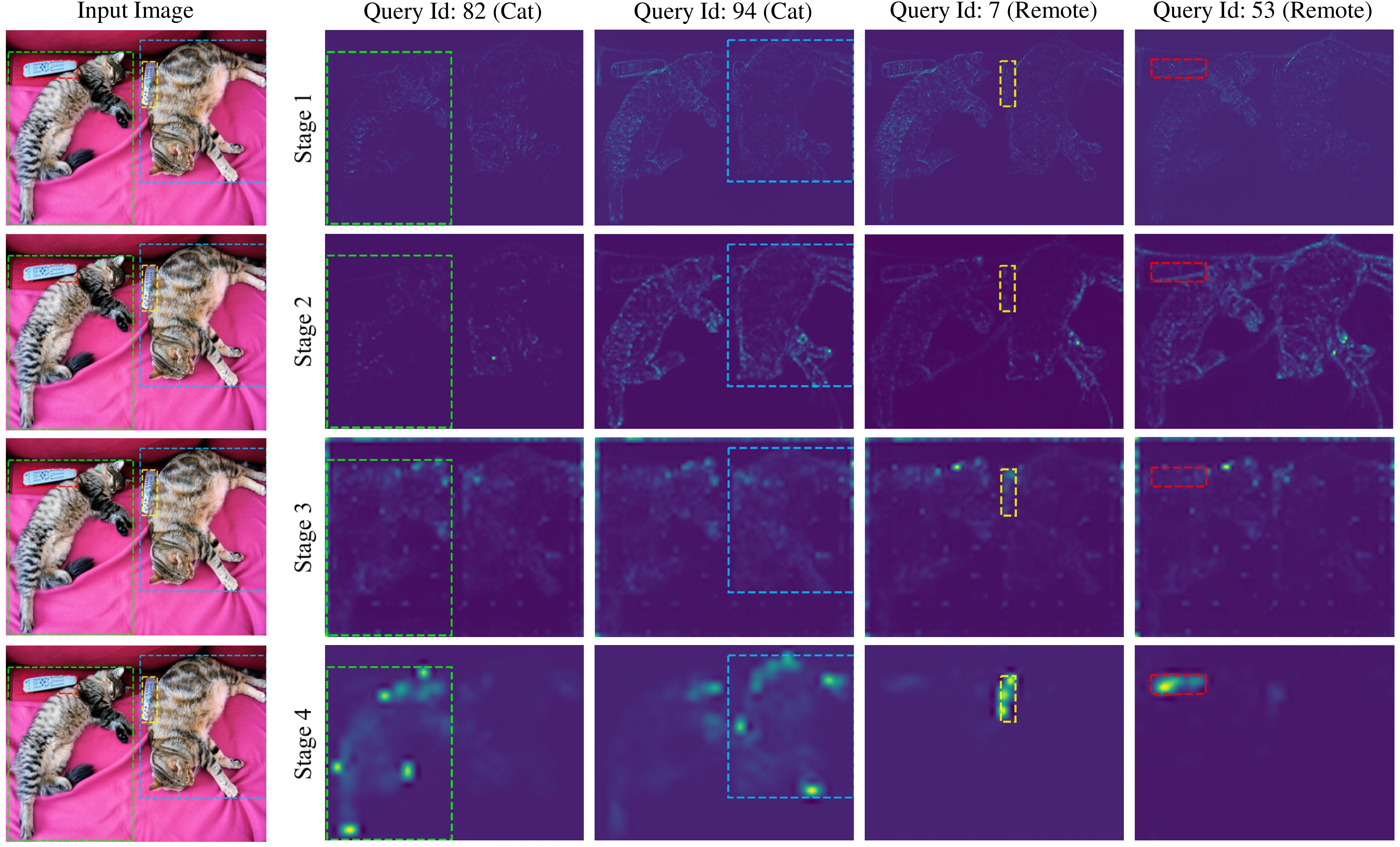}
\end{center}
\vspace*{-0.25cm}
\hspace*{6.3cm}{\small (a) Cross-attention at \textbf{all} stages $\{1, 2, 3, 4\}$.}
\vspace*{-0.3cm}
\begin{center}
\includegraphics[height=30.25mm]{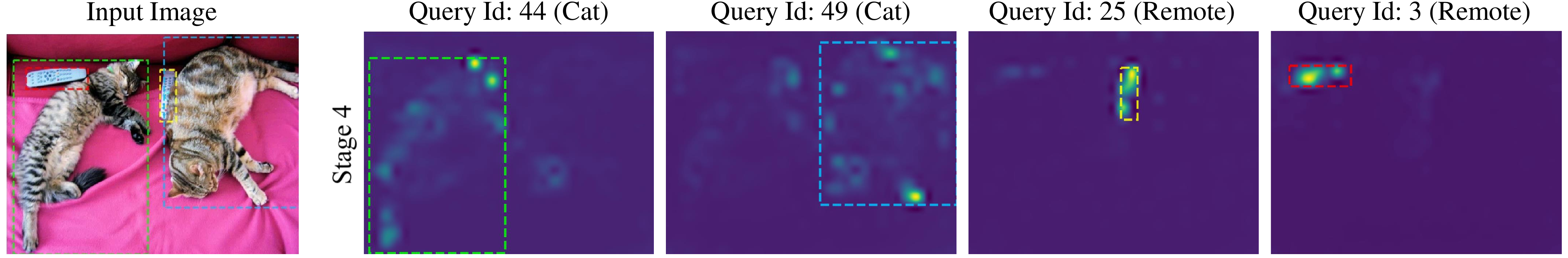}
\end{center}
\vspace*{-0.15cm}
\hspace*{6.5cm}{\small (b) Cross-attention at the \textbf{last} stage $\{4\}$.}
\vspace*{-0.2cm}
\caption{{Visualization of the attention map for cross-attention with \algname\,(Swin-nano).}}
\label{fig:attention_analysis}
\vspace*{-0.40cm}
\label{fig:attention}
\end{figure*}

\smallskip
{
\noindent{\textbf{Deformable DETR}} shows significant improvement in FPS\,($+14.4$) but its AP drops sharply ($-9.1$) when its neck encoder is removed. 
%
%
Thus, it is difficult to obtain fine-grained object detection representation directly from the raw ViT backbone without using an additional neck encoder. 
However, ViDT compensates for the effect of the neck encoder by adding $[\mathtt{DET}]$ tokens into the body (backbone), thus successfully removing the computational bottleneck without compromising AP; it maintains 6.4 higher AP compared with the neck encoder-free Deformable DETR (the second row) while achieving similar FPS. 
%
This can be attributed to that RAM has a great contribution to the performance w.r.t AP and FPS, especially for the trade-off between them. 
}

\smallskip
\noindent\textbf{YOLOS} shows a significant gain in AP\,($+7.7)$ while losing FPS\,($-11.0$) when the neck decoder is added. Unlike Deformable DETR, its AP significantly increases even without the neck encoder due to the use of a standalone object detector as its backbone\,(i.e., the modified DeiT in Figure \ref{fig:pipelines}(b)). However, its AP is lower than \algname{} by 2.3AP. Even worse, it is not scalable for large models because of its quadratic computational cost for attention. Therefore, in the aspects of accuracy and speed, \algname{} maintains its dominance compared with the two carefully tuned baselines.

For a complete analysis, we additionally add a neck encoder to \algname{}. The inference speed of \algname{} degrades drastically by $13.7$ because of the self-attention for multi-scale features at the neck encoder. However, it is interesting to see the improvement of AP by $5.7$ while adding only 3M parameters; it is $3.0$ higher even than Deformable DETR. This indicates that lowering the computational complexity of the encoder and thus increasing its utilization could be another possible direction for a fully transformer-based object detector.

\subsection{$[\mathtt{DET}]\times[\mathtt{PATCH}]$ Attention in RAM}
\label{appendix:analysis_attention}

In Section \ref{sec:exp_ram}, it turns out that the cross-attention in RAM is only necessary at the last stage of Swin Transformer; all the different selective strategies show similar AP as long as cross-attention is activated at the last stage. Hence, we analyze the attention map obtained by the cross-attention in RAM. Figure \ref{fig:attention_analysis} shows attention maps for the stages of Swin Transformer where cross-attention is utilized; it contrasts (a)\,\algname{} with cross-attention at all stages and (b)\,\algname{} with cross-attention at the last stage. Regardless of the use of cross-attention at the lower stage, it is noteworthy that the finally obtained attention map at the last stage is almost the same. In particular, the attention map at Stage 1--3 does not properly focus the features on the target object, which is framed by the bounding box. In addition, the attention weights\,(color intensity) at Stage 1--3 are much lower than those at Stage 4.  Since features are extracted from a low level to a high level in a bottom-up manner as they go through the stages, it seems difficult to directly get information about the target object with such low-level features at the lower level of stages. Therefore, this analysis provides strong empirical evidence for the use of selective $[\mathtt{DET}]\times[\mathtt{PATH}]$ cross-attention.

{
\subsection{$[\mathtt{DET}]\times[\mathtt{DET}]$ Attention in RAM}
\label{appendix:comp_det}
}

\begin{table}
\caption{{AP and FPS comparison with different $[\mathtt{DET}]\times[\mathtt{DET}]$ self-attention strategies with ViDT.}}
\vspace*{-0.3cm}
\label{table:ablation_det}
\vspace*{0.2cm}
\small
\begin{tabular}{L{0.5cm} |X{0.7cm} X{0.7cm} X{0.7cm} X{0.7cm} |X{1.3cm} X{1.3cm}  }\hline
\rowcolor{Gray}
    &   \multicolumn{4}{c|}{Stage Id}  & \multicolumn{2}{c}{Swin-nano}\\
\rowcolor{Gray}
\!~\#\!  & \!\!\!1\!\!\! & \!\!\!2\!\!\! & \!\!\!3\!\!\! & \!\!\!4\!\!\!  & \!\!AP\!\! &  \!\!FPS\!\! \\\toprule
\!(1)\!\! & \!\!\checkmark\!\! & \!\!\checkmark\!\! & \!\!\checkmark\!\! & \!\!\checkmark\!\! & \!\! {40.4}\!\! & \!\! {40.8 (76.0)}\!\!  \\
\!(2)\!\! & & \!\!\checkmark\!\! & \!\!\checkmark\!\! & \!\!\checkmark\!\! & \!\! {40.3}\!\! & \!\! {41.0 (76.4)}\!\!  \\
\!(3)\!\! & & & \!\!\checkmark\!\! & \!\!\checkmark\!\! & \!\! {40.4}\!\! & \!\! {41.1 (77.0)}\!\!  \\
\!(4)\!\! & & & & \!\!\checkmark\!\! & \!\! {40.1}\!\! & \!\! {41.5 (78.1)}\!\!  \\
\!(5)\!\! & & & & & \!\! {39.7}\!\! & \!\! {41.7 (78.5)}\!\!  \\\bottomrule
\end{tabular}
\vspace*{-0.3cm}
\end{table}

Another possible consideration for \algname{} is the use of $[\mathtt{DET}]\times[\mathtt{DET}]$ self-attention in RAM.
{We conduct an ablation study by removing the $[\mathtt{DET}]\times[\mathtt{DET}]$ attention one by one from the bottom stage, and summarize the results in Table \ref{table:ablation_det}. When all the $[\mathtt{DET}]\times[\mathtt{DET}]$ self-attention are removed, (5) the AP drops by 0.7, which is a meaningful performance degradation. On the other hand, as long as the self-attention is activated at the last two stages, (1) -- (3) all the strategies exhibit similar AP. Therefore, only keeping $[\mathtt{DET}]\times[\mathtt{DET}]$ self-attention at the last two stages can further increase FPS ($+0.3$) without degradation in AP. This observation could be used as another design choice for the AP and FPS trade-off. Therefore, we believe that $[\mathtt{DET}]\times[\mathtt{DET}]$ self-attention is meaningful to use in RAM.}

\section{Preliminaries: Transformers}
\label{appendix:transformer}

A transformer is a deep model that entirely relies on the self-attention mechanism for machine translation\,\cite{vaswani2017attention}. In this section, we briefly revisit the standard form of the transformer.

\smallskip
\noindent\textbf{Single-head Attention.}
The basic building block of the transformer is a self-attention module, which generates a weighted sum of the values\,(contents), where the weight assigned to each value is the attention score computed by the scaled dot-product between its query and key. Let $W_{Q}$, $W_{K}$, and $W_{V}$ be the learned projection matrices of the attention module, and then the output is generated by
\begin{equation}
\begin{gathered}
\text{Attn}(Z) =\text{softmax}\big(\frac{(ZW_{Q})(ZW_{K})^{\top}}{\sqrt{d}}\big)(ZW_{V}) \in \mathbb{R}^{hw \times d}, \\
\text{where} ~~W_{Q}, W_{K}, W_{V} \in \mathbb{R}^{d \times d}.
\end{gathered}
\end{equation}

\noindent\textbf{Multi-head Attention.}
It is beneficial to maintain multiple heads such that they repeat the linear projection process $k$ times with different learned projection matrices. Let $W_{Q_i}$, $W_{K_i}$, and $W_{V_i}$ be the learned projection matrices of the $i$-th attention head. Then, the output is generated by the concatenation of the results from all heads,
\begin{equation}
\begin{gathered}
\text{Multi-Head}(Z) =\\
[\text{Attn}_{1}(Z), \text{Attn}_{2}(Z), \dots, \text{Attn}_{k}(Z)] \in \mathbb{R}^{hw \times d}, \\
\text{where} ~~\forall_{i}~W_{Q_i}, W_{K_i}, W_{V_i} \in \mathbb{R}^{d \times (d/k)}.
\end{gathered}
\end{equation}
Typically, the dimension of each head is divided by the total number of heads.

\noindent \textbf{Feed-Forward Networks\,(FFNs).} The output of the multi-head attention is fed to the point-wise FFNs, which performs the linear transformation for each position separately and identically to allow the model focusing on the contents of different representation subspaces. Here, the residual connection and layer normalization are applied before and after the FFNs. The final output is generated by \looseness=-1
\begin{equation}
\begin{gathered}
H= \text{LayerNorm}(\text{Dropout}(H^\prime)+ H^{\prime\prime}),\\
\text{where}~~ H^\prime~ = \text{FFN}(H^{\prime\prime})~~ \text{and}\\ H^{\prime\prime} = \text{LayerNorm}(\text{Dropout}(\text{Multi-Head}(Z))+Z).
\end{gathered}
\end{equation}

\noindent\textbf{Multi-Layer Transformers.} The output of a previous layer is fed directly to the input of the next layer. Regarding the positional encoding, the same value is added to the input of each attention module for all layers.